\pdfoutput=1

\documentclass[11pt]{article}

\usepackage[final]{acl}

\usepackage{times}
\usepackage{latexsym}
\usepackage{cleveref}
\usepackage{CJKutf8}
\usepackage{longtable}

\usepackage[T1]{fontenc}

\usepackage[utf8]{inputenc}

\usepackage{microtype}

\usepackage{inconsolata}

\usepackage{graphicx}
\usepackage{subcaption}
\usepackage{float}
\usepackage{afterpage}
\usepackage{xcolor}    
\usepackage{tcolorbox} 
\usepackage{soul}
\sethlcolor{red!30} 
\usepackage{fontawesome5}

\title{M-IFEval: Multilingual Instruction-Following Evaluation}

\author{
  \textbf{Antoine Dussolle\textsuperscript{1,2\thanks{Work done at Lightblue KK.}}},
  \textbf{Andrea Cardeña Díaz\textsuperscript{1}},
  \textbf{Shota Sato\textsuperscript{1}},
  \textbf{Peter Devine\textsuperscript{1}},
\\
\\
  \textsuperscript{1}Lightblue KK.,
  \textsuperscript{2}Arkema,
\\
    \texttt{antoine.dussolle@gmail.com}
    \\
    \texttt{andrea.cdiaz@hotmail.es}
    \\
    \texttt{\{shota.sato,peter\}@lightblue-tech.com}
}   

\begin{document}
\maketitle
\begin{abstract}

Instruction following is a core capability of modern Large language models (LLMs), making evaluating this capability essential to understanding these models. The Instruction Following Evaluation (IFEval) benchmark from the literature does this using objective criteria, offering a measure of LLM performance without subjective AI or human judgement. However, it only includes English instructions, limiting its ability to assess LLMs in other languages. 

We propose the Multilingual Instruction Following Evaluation (M-IFEval) benchmark, expanding the evaluation to French, Japanese, and Spanish, with both general and language-specific instructions. 
Applying this benchmark to 8 state-of-the-art LLMs, we find that benchmark performance across languages and instruction types can vary widely, underscoring the importance of a multilingual benchmark for evaluating LLMs in a diverse cultural context.

\end{abstract}

\section{Introduction}


Large language models (LLMs) have demonstrated amazing accuracy in many fields including medicine~\cite{tian-etal-2024-chimed,frisoni-etal-2024-generate}, law~\cite{jiang-etal-2024-leveraging}, and education~\cite{luo-etal-2024-chain}. However, their accuracy has also shown to be low for some tasks such as reasoning~\cite{tong-etal-2024-llms} and cultural understanding~\cite{wang-etal-2024-countries}.

One type of task of particular importance for LLMs is that of instruction following, where an LLM must carry out the instructions of the user in a ``zero-shot'' setting (i.e. without necessarily being trained specifically to perform that instruction)~\cite{zhong2021adapting,mishra2022cross,weifinetuned,sanh2022multitask}.

Benchmarks such as Instruction-Following Evaluation (IFEval)~\cite{zhou2023instruction} have proposed ways of evaluating LLMs on instruction following without the need for using an external LLM-as-a-judge~\cite{zheng2023judging}, which may exhibit self-enhancement bias~\cite{xu-etal-2024-pride}.

However, this benchmark is a purely English-based benchmark, raising questions as to the applicability of its results to other languages. While some efforts have been made to make a multilingual version of this benchmark, at present this only extends as far as translating the original prompts into other languages~\cite{qwen2.5}.
This approach fails to evaluate aspects of instruction following that are specific to different languages. Specifically developed code is required to understand whether an LLM can, for example, uses the correct punctuation or script for a given language when prompted.

We present Multilingual Instruction Following Evaluation (M-IFEval), a benchmark for evaluating LLM instruction following beyond English.
Our benchmark consists of three popular natural languages, French, Japanese, and Spanish, and contains both instructions previously assessed in English as well as novel instructions that are specific to our evaluation languages. We assess 8 state-of-the-art LLMs using M-IFEval and compare their evaluation results to the original English IFEval scores.

Our evaluation results show that, among the models tested on the English instruction-following benchmark, widely-used LLMs like GPT4o achieve the highest relative performance. However, for benchmarks in other languages, models such as o1 and Sonnet perform better in instruction following. We also highlight that state-of-the-art LLMs achieve surprisingly low scores on some language-specific instructions such as using or not using special characters or scripts.

Our work demonstrates the value of a multilingual benchmark when selecting LLMs for a non-English based task and identifies key areas for improvement, such as character- and script-level instructions,  in modern LLMs. We make the evaluation code and data for this benchmark available online\footnote{\url{https://github.com/lightblue-tech/M-IFEval}}.

\section{Related work}


Benchmarks like GLUE~\cite{wang-etal-2018-glue}, ARC~\cite{allenai:arc}, SuperGLUE~\cite{wang2019superglue}, Winogrande~\cite{sakaguchi2019winogrande}, HellaSwag~\cite{zellers2019hellaswag}, MMLU~\cite{hendryckstest2021}, and others, that contain objective tasks such as natural language inference and semantic similarity have been widely used within the literature to evaluate LLMs~\cite{anil2023palm,le2023bloom,dettmers2023qlora}. However, these tasks do not fully represent the realistic usage of LLMs in practical scenarios, for example as conversational agents or decision-making systems.

Other benchmarks such as MT-Bench~\cite{zheng2023judging}, AlpacaEval~\cite{dubois2024length}, Arena-Hard~\cite{li2024crowdsourced} and InFoBench~\cite{qin-etal-2024-infobench} provide a framework for evaluating LLMs in a more practical conversational or instruction-based setting. However, their reliance on subjective AI scoring raises self-enhancement bias concerns~\cite{xu-etal-2024-pride}, making them less suitable for model evaluation.

Chatbot Arena~\cite{zheng2023judging} involves evaluating LLMs using human users, thus removing the potential for self-enhancement bias. However, Chatbot Arena requires large-scale deployment, making it difficult to replicate this evaluation for local models.

IFEval~\cite{zhou2023instruction} was designed to evaluate the ability of LLMs to follow instructions in more practical scenarios, making it an effective measure of models intended for real-world usage. Unlike MT-Bench, IFEval uses objective criteria for evaluation, removing the subjectivity inherent in AI-judged systems. However, its primary limitation is that it is currently available only in English, which restricts its applicability for evaluating multilingual models.


The model evaluation of the instruction-tuned Qwen 2.5~\cite{qwen2.5} extended IFEval to support multilingual settings by translating 100 examples per language and removing instructions that were not applicable to a given language. While this approach allows for deterministic evaluation of LLMs on multilingual data, it may neglect language-specific aspects of instruction-following.

To overcome the limitations of existing benchmarks, we propose a new, multilingual version of IFEval that is not simply a translation of previous datasets but contains language-specific instructions for novel evaluation.


\section{Method}

This section details how we constructed the M-IFEval benchmark and how we used it to evaluate various state-of-the-art LLMs.

We first chose the languages that would be included in our benchmark. We chose French, Japanese, and Spanish as our research team included one native speaker in each language, with each acting as language lead for their respective language.

Our team consulted the list of instructions included in the original English language IFEval benchmark and considered any that were not applicable to their language. Following this, we removed the word number length constraint instruction and the case change instructions (all uppercase, all lowercase, and frequency of all capital words) for Japanese, as its writing system does not account for letter case.

Next, each language lead created a list of instructions that could be evaluated using objective criteria, including those specific to their language. The design of these verifiable instructions was guided by two key considerations. First, the instructions needed to evaluate language-specific linguistic and textual control, addressing elements such as diacritics, script-level constraints, and cultural nuances. Secondly, they were designed to maintain a reasonable difficulty level, ensuring that the tasks remained easily achievable for native speakers, thus promoting fairness across the three evaluation languages. These instructions encompassed grammatical, stylistic, and script-based elements tailored to each language.

The \textbf{Spanish-specific instructions} consisted of three special character-based instructions (ñ frequency, accent frequency, and ü frequency) and two punctuation-based instructions (using grammatically correct question marks and exclamation marks).

The \textbf{French-specific instructions} consisted of three special character-based instructions (forbidding use of œ/ç, forbidding accents, and adding the correct accents to a given text), and two content based instructions (not using Arabic numerals in the response and using the informal direct way of addressing someone).

\begin{CJK}{UTF8}{min}
The \textbf{Japanese-specific instructions} consisted of seven script based instructions (only/do not use katakana, only/do not use hiragana, use at least/most N kanji, include furigana, write all numbers as kanji), two format based instructions (responses must be a numbered list of N items, responses must include at least N taigen-dome\footnote{A Japanese grammatical structure where a sentence ends with a noun or noun phrase~\cite{hayashi2007sentence}}), and one instruction each of a length based instruction (use at least/most N characters), a start/end based instruction (end all sentences with です/ます), and a punctuation based instruction (do not use 。 - a Japanese period).
\end{CJK}

The list of all language-specific instructions can be found in \cref{tab:all_instr} in the appendix.

This resulted in a list of instructions for each language which were a mix of instructions from the original IFEval benchmark and instructions that were specific to their language. 

Each language lead then developed a function for each instruction that evaluates whether a response did or did not correctly follow the given instruction. These functions were then added to the evaluation codebase of the original IFEval benchmark.

Prompts that instruct the LLM to follow at least one instruction were then developed by the language leads in a similar way to the original IFEval work. Prompts were developed by generating multiple example prompts using a state-of-the-art LLM before selecting and editing prompts manually to obtain a list of multiple prompts per instruction. As with the original IFEval, we constructed prompts with one, two, and three instructions contained within the same prompt.
The correct evaluation arguments (e.g. specifying \verb|2| if the prompt specifies 2 sentences in the sentence counting instruction) were then manually added to each instruction.

This process resulted in 115, 172, and 235 prompts for Spanish, Japanese, and French, respectively, with at least 4 unique prompts per instruction for Spanish and Japanese, and 7 prompts per instruction for French. For Spanish, Japanese, and French, our benchmark contains 8, 34, and 68 prompts respectively that contains 2 instructions, and 7, 10, and 21 prompts that contain 3 instructions. For a clearer understanding of the dataset's structure, we provide additional details in \cref{Appendix: dataset stat}, including basic statistics, the number of prompts per instruction, and the distribution of prompts across instruction groups by language.

Responses to these prompts were then generated using all the state-of-the-art LLMs that we had access to, consisting of 4 versions of OpenAI's GPT (GPT4o, GPT4o Mini, o1, o1 Mini), 3 versions of Anthropic's Claude 3.5 (Opus, Sonnet, and Haiku), and the largest multilingual open source LLM that we could run in 4 bits on a single 40GB A100 GPU (Qwen 2.5 32B Instruct GPTQ Int4~\cite{qwen2.5}). Whenever possible, responses were generated using greedy decoding (temperature set to 0) to ensure reproducibility.

These responses were then evaluated using our evaluation code and we report the average score in each language for each model. We separately report the scores only of the average language-specific instructions for each language and model.
As with the original IFEval work, we calculate both the strict and loose scores for each instruction. We report the strict scores in the main document and report the loose scores in the appendix.



\section{Results}

\begin{table}[]
\centering
\resizebox{\columnwidth}{!}{
\begin{tabular}{|c|r|rrr|r}
\cline{1-5}
\multicolumn{1}{|c|}{\textbf{Model name}} & \textbf{EN} & \textbf{ES} & \textbf{FR} & \textbf{JA} & \textbf{Mean} \\ \cline{1-5}
o1\textsuperscript{$\dag$} & 86.7 & \textbf{92.7} & \textbf{91.3} & 75.7 & \textbf{86.6} \\
Opus\textsuperscript{$\ddag$} & 87.3 & 90.5 & 87.0 & 75.7 & 84.4 \\
Sonnet\textsuperscript{$\ddag$} & 88.1 & 87.6 & 88.1 & \textbf{77.0} & 84.2 \\
o1 Mini\textsuperscript{$\dag$} & 83.9 & 92.0 & 88.4 & 69.5 & 83.3 \\
GPT4o\textsuperscript{$\dag$} & \textbf{88.6} & 89.8 & 87.8 & 70.4 & 82.7 \\
GPT4o Mini\textsuperscript{$\dag$} & 86.0 & 85.4 & 85.5 & 65.9 & 78.9 \\
Qwen 2.5 32B I.\textsuperscript{$\star$} & 86.0 & 82.5 & 81.7 & 65.9 & 76.7 \\
Haiku\textsuperscript{$\ddag$} & 77.3 & 78.8 & 78.3 & 61.9 & 73.0 \\ \cline{1-5}
\end{tabular}
}
\caption{Average strict scores of M-IFEval for each language for each model evaluated, sorted by the mean combined Spanish, French, and Japanese scores.}
\label{tab:avg_scores}
\end{table}

\begin{table}[]
\centering
\resizebox{\columnwidth}{!}{
\begin{tabular}{|c|rrr|r}
\cline{1-4}
\multicolumn{1}{|c|}{\textbf{Model name}} & \textbf{ES} & \textbf{FR} & \textbf{JA} & \textbf{Mean} \\ \cline{1-4}
o1\textsuperscript{$\dag$} & \textbf{75.0} & \textbf{96.1} & 61.4 & \textbf{77.5} \\
Sonnet\textsuperscript{$\ddag$} & 66.7 & 90.2 & \textbf{70.5} & 75.8 \\
Opus\textsuperscript{$\ddag$} & 62.5 & 90.2 & 64.8 & 72.5 \\
GPT4o\textsuperscript{$\dag$} & 58.3 & 80.4 & 55.7 & 64.8 \\
o1 Mini\textsuperscript{$\dag$} & 66.7 & 72.5 & 50.0 & 63.1 \\
Qwen 2.5 32B I.\textsuperscript{$\star$} & 54.2 & 78.4 & 54.5 & 62.4 \\
Haiku\textsuperscript{$\ddag$} & 54.2 & 80.4 & 52.3 & 62.3 \\
GPT4o Mini\textsuperscript{$\dag$} & 58.3 & 68.6 & 47.7 & 58.2 \\ \cline{1-4}
\end{tabular}
}
\caption{Average strict scores of M-IFEval for each language only on the instructions that are specific to that language, sorted by the mean combined Spanish, French, and Japanese scores. \\\textit{\textsuperscript{$\dag$}OpenAI, \textsuperscript{$\ddag$}Anthropic,\textsuperscript{$\star$}Qwen}}
\label{tab:avg_unique_scores}
\end{table}

\Cref{tab:avg_scores} shows the average evaluation score for each model evaluated in each language in the M-IFEval benchmark, along with with the English scores in the original IFEval benchmark. 

We observe that while GPT4o and Sonnet are the top two models for English M-IFEval, o1 and Opus have the highest score on average for the three languages in our benchmark. We also observe a greater spread in scores between the best and worse performing models in our evaluation compared to the English IFEval, with the best and worst scores on the original English benchmark having a difference of 11.3 percentage points, while we observe differences of 13.9, 13.0, and 15.1 for Spanish, French, and Japanese respectively.

\Cref{tab:avg_unique_scores} shows the average scores only on instructions that are unique to each language. We observe that while the o1 model attains the greatest scores on Spanish and French benchmarks, Sonnet achieves markedly higher scores on the Japanese benchmark. When we analysed the scores only of instructions that had been included in the original IFEval benchmark (i.e. instructions not unique to the language), we found that o1 achieves a score on the Japanese benchmark of 84.8 while Sonnet achieves a score of 81.2.


When we analysed specific instructions with the lowest average evaluation scores across all models that we tested, we found that the 10 instructions with the lowest scores all were language-specific instructions such as forbidding ``œ/ç'', forbidding katakana, or specifying the frequency of the ``ñ'' character. The average scores for these three instruction types across all models was 60.2\%, 14.3\%, and 0.0\%, respectively. Conversely, we observe that LLMs attain high scores in following instructions such as adding accents to French text, adding Spanish question marks/exclamantion marks, and making both French and Spanish text uppercase/lowercase. This suggests that while many LLMs perform well in following formatting instructions, such as structuring outputs and arranging punctuation, they struggle with script-based instructions. This is apparent from the drop in accuracy in the `Special character' instruction group for Spanish and French, as well as the `Script' instruction group for Japanese, both of which mostly comprise character-level instructions, as shown in  figures \ref{fig:es_performance}--\ref{fig:ja_performance} in \cref{Appendix: detailed results}.

The full scores averaged across all models for each instruction can be found in \cref{tab:all_instr_averaged}.

\section{Discussion \& Future work}

Overall, our results show that modern LLMs are generally proficient at instruction following outside of English. However, our evaluation scores still vary between both languages and task types, indicating the need for future improvement of LLMs in a wide range of linguistically and culturally important tasks.

Our results show that Sonnet is more proficient at Japanese-specific instructions compared to o1, whereas o1 is more proficient at Spanish and French specific instructions.
This could indicate that o1 has been trained on more Spanish and French data, or linguistically similar languages that confer cross lingual generalisation~\cite{snaebjarnarson-etal-2023-transfer,muennighoff2023crosslingual}, while Sonnet may have been trained on more Japanese data.

Moreover, we find that the highest performing model in our English evaluation was neither o1 nor Sonnet, but GPT4o. This highlights the need for multilingual LLM evaluations to select the best model for a target language, as no single LLM excels in all languages.

Our results also show that LLMs generally achieve poor performance on seemingly simple language-specific tasks such as restricting usage of a given script (e.g. ``write your answer without using any katakana'') or controlling for the amount of times a certain special character is used (e.g. ``use the `ñ' character exactly 5 times in your response''). Examples of such failures are provided in \cref{Appendix: failure examples}. This contrasts with high English scores for similar tasks (e.g. ``use the letter c at least 60 times in your response''). This may reflect a gap between LLM performance in English to that of other languages which has been observed in other tasks~\cite{ahuja2024megaverse,10.1145/3589334.3645643}.

Future work could consider exactly why the performance of LLMs varies for different languages.
Previous work has investigated the effect of different language mixtures on downstream tasks~\cite{ustun-etal-2024-aya,wei2023polylm}, so experiments involving different mixes of multilingual pre-training data and fine-tuning data could possibly show the effect of training data on instruction following performance.

Experiments using a byte-level tokenizer~\cite{xue-etal-2022-byt5} could possibly answer the question of why script or character based instructions are so hard to follow for modern token-level LLMs.



\section{Conclusion}

In this paper, we have presented M-IFEval, a multilingual benchmark which evaluates the instruction following abilities of LLMs in three non-English languages: French, Japanese, and Spanish.

Our results show that while GPT4o achieves the greatest instruction following performance on the English IFEval benchmark, we find that other models, o1 and Sonnet, achieve higher scores on M-IFEval.

This finding highlights the importance of multilingual evaluation in assessing a model's instruction following abilities.

We also identify several types of instructions for which the average LLM performance was surprisingly low. This includes specifying the usage/non usage of a certain script and specifying the frequency of a certain amount of non-English characters.

This work contributes a new benchmark to the field of multilingual evaluation of LLMs and provides observations for what these models can and cannot do in the context of multilingual instruction following.

\section{Limitations}

One of the limitations of this work is that our benchmark only considers instructions that can be objectively evaluated using simple string checking code. This means that our evaluation does not include any of the large group of possible instructions which would require more intricate analysis to evaluate upon (e.g. translation quality, fact checking, question answering). We acknowledge this and leave more detailed evaluation of LLMs on specific tasks to other benchmarks such as XNLI~\cite{conneau2018xnli}, XQuad~\cite{artetxe2020cross}, and Flores~\cite{costa2022no}. And, although certain instructions that can be evaluated programmatically might seem unnatural (e.g., "Write a paragraph using the letter 'j' exactly 9 times"), our goal was to investigate the types of instructions that LLMs still tend to struggle with the most. This therefore provides insight into the types of realistic tasks these LLMs may also find challenging. Future work could explore instruction following in more realistic, user-driven scenarios by incorporating organic, diverse, and contextually grounded prompts that better reflect real-world usage. This would provide a more nuanced understanding of how well models perform in genuinely practical settings.

Another limitation of this work is that we only consider three non-English languages in our evaluation. Moreover, these three languages were all relatively high-resource languages, and since we observe a gap between English and our evaluation languages, we may observe an even greater gap for low resource languages. Future work could include adding more languages to our benchmark, particularly low resource languages. This could entail adding more language-specific instructions (e.g. converting ``Boko'', or Latin, script in Hausa to ``Ajami'', or Arabic, script~\cite{abdulmumin2014survey}) to further identify if there are any other tasks in which LLMs perform particularly poorly.

A final limitation of this work is that we only evaluate over 8 state-of-the-art LLMs in our evaluation when other LLMs such as Gemini~\cite{reid2024gemini} are also available. This was done due to a combination of technical, financial, and document-space limitations, and so the main contributions of our paper are that we demonstrate that relative instruction following performance is not uniform across all languages for a given LLM, and that some of the top performing LLMs available still cannot perform basic tasks such as controlling special character usage. We leave it for future work to use this benchmark to compare their models against others.





\bibliography{custom}

\begin{thebibliography}{39}
\providecommand{\natexlab}[1]{#1}

\bibitem[{Abdulmumin(2014)}]{abdulmumin2014survey}
SA~Abdulmumin. 2014.
\newblock A survey of historical prevalence of hausa language in contemporary literacy.
\newblock \emph{ZAHIRA--Journal of Historical Research, Dept. of History, Ahmadu Bello University, ZARIA Nigeria}, 5(4).

\bibitem[{Ahuja et~al.(2024)Ahuja, Aggarwal, Gumma, Watts, Sathe, Ochieng, Hada, Jain, Ahmed, Bali et~al.}]{ahuja2024megaverse}
Sanchit Ahuja, Divyanshu Aggarwal, Varun Gumma, Ishaan Watts, Ashutosh Sathe, Millicent Ochieng, Rishav Hada, Prachi Jain, Mohamed Ahmed, Kalika Bali, et~al. 2024.
\newblock Megaverse: Benchmarking large language models across languages, modalities, models and tasks.
\newblock In \emph{Proceedings of the 2024 Conference of the North American Chapter of the Association for Computational Linguistics: Human Language Technologies (Volume 1: Long Papers)}, pages 2598--2637.

\bibitem[{Anil et~al.(2023)Anil, Dai, Firat, Johnson, Lepikhin, Passos, Shakeri, Taropa, Bailey, Chen et~al.}]{anil2023palm}
Rohan Anil, Andrew~M Dai, Orhan Firat, Melvin Johnson, Dmitry Lepikhin, Alexandre Passos, Siamak Shakeri, Emanuel Taropa, Paige Bailey, Zhifeng Chen, et~al. 2023.
\newblock Palm 2 technical report.
\newblock \emph{arXiv preprint arXiv:2305.10403}.

\bibitem[{Artetxe et~al.(2020)Artetxe, Ruder, and Yogatama}]{artetxe2020cross}
Mikel Artetxe, Sebastian Ruder, and Dani Yogatama. 2020.
\newblock On the cross-lingual transferability of monolingual representations.
\newblock In \emph{Proceedings of the 58th Annual Meeting of the Association for Computational Linguistics}. Association for Computational Linguistics.

\bibitem[{Clark et~al.(2018)Clark, Cowhey, Etzioni, Khot, Sabharwal, Schoenick, and Tafjord}]{allenai:arc}
Peter Clark, Isaac Cowhey, Oren Etzioni, Tushar Khot, Ashish Sabharwal, Carissa Schoenick, and Oyvind Tafjord. 2018.
\newblock Think you have solved question answering? try arc, the ai2 reasoning challenge.
\newblock \emph{arXiv:1803.05457v1}.

\bibitem[{Conneau et~al.(2018)Conneau, Rinott, Lample, Williams, Bowman, Schwenk, and Stoyanov}]{conneau2018xnli}
Alexis Conneau, Ruty Rinott, Guillaume Lample, Adina Williams, Samuel Bowman, Holger Schwenk, and Veselin Stoyanov. 2018.
\newblock Xnli: Evaluating cross-lingual sentence representations.
\newblock In \emph{Proceedings of the 2018 Conference on Empirical Methods in Natural Language Processing}, pages 2475--2485.

\bibitem[{Costa-juss{\`a} et~al.(2022)Costa-juss{\`a}, Cross, {\c{C}}elebi, Elbayad, Heafield, Heffernan, Kalbassi, Lam, Licht, Maillard et~al.}]{costa2022no}
Marta~R Costa-juss{\`a}, James Cross, Onur {\c{C}}elebi, Maha Elbayad, Kenneth Heafield, Kevin Heffernan, Elahe Kalbassi, Janice Lam, Daniel Licht, Jean Maillard, et~al. 2022.
\newblock No language left behind: Scaling human-centered machine translation.
\newblock \emph{arXiv preprint arXiv:2207.04672}.

\bibitem[{Dettmers et~al.(2023)Dettmers, Pagnoni, Holtzman, and Zettlemoyer}]{dettmers2023qlora}
Tim Dettmers, Artidoro Pagnoni, Ari Holtzman, and Luke Zettlemoyer. 2023.
\newblock Qlora: Efficient finetuning of quantized llms.
\newblock \emph{arXiv preprint arXiv:2305.14314}.

\bibitem[{Dubois et~al.(2024)Dubois, Galambosi, Liang, and Hashimoto}]{dubois2024length}
Yann Dubois, Bal{\'a}zs Galambosi, Percy Liang, and Tatsunori~B Hashimoto. 2024.
\newblock Length-controlled alpacaeval: A simple way to debias automatic evaluators.
\newblock \emph{arXiv preprint arXiv:2404.04475}.

\bibitem[{Frisoni et~al.(2024)Frisoni, Cocchieri, Presepi, Moro, and Meng}]{frisoni-etal-2024-generate}
Giacomo Frisoni, Alessio Cocchieri, Alex Presepi, Gianluca Moro, and Zaiqiao Meng. 2024.
\newblock \href {https://doi.org/10.18653/v1/2024.acl-long.533} {To generate or to retrieve? on the effectiveness of artificial contexts for medical open-domain question answering}.
\newblock In \emph{Proceedings of the 62nd Annual Meeting of the Association for Computational Linguistics (Volume 1: Long Papers)}, pages 9878--9919, Bangkok, Thailand. Association for Computational Linguistics.

\bibitem[{Hayashi and Matsubara(2007)}]{hayashi2007sentence}
Yukiko Hayashi and Shigeki Matsubara. 2007.
\newblock Sentence-style conversion of japanese news article for text-to-speech application.
\newblock In \emph{Proceedings of 7th International Symposium on Natural Language Processing}, pages 257--262.

\bibitem[{Hendrycks et~al.(2021)Hendrycks, Burns, Basart, Zou, Mazeika, Song, and Steinhardt}]{hendryckstest2021}
Dan Hendrycks, Collin Burns, Steven Basart, Andy Zou, Mantas Mazeika, Dawn Song, and Jacob Steinhardt. 2021.
\newblock Measuring massive multitask language understanding.
\newblock \emph{Proceedings of the International Conference on Learning Representations (ICLR)}.

\bibitem[{Jiang et~al.(2024)Jiang, Zhang, Mahari, Kessler, Ma, August, Li, Pentland, Kim, Roy, and Kabbara}]{jiang-etal-2024-leveraging}
Hang Jiang, Xiajie Zhang, Robert Mahari, Daniel Kessler, Eric Ma, Tal August, Irene Li, Alex Pentland, Yoon Kim, Deb Roy, and Jad Kabbara. 2024.
\newblock \href {https://doi.org/10.18653/v1/2024.acl-long.388} {Leveraging large language models for learning complex legal concepts through storytelling}.
\newblock In \emph{Proceedings of the 62nd Annual Meeting of the Association for Computational Linguistics (Volume 1: Long Papers)}, pages 7194--7219, Bangkok, Thailand. Association for Computational Linguistics.

\bibitem[{Jin et~al.(2024)Jin, Chandra, Verma, Hu, De~Choudhury, and Kumar}]{10.1145/3589334.3645643}
Yiqiao Jin, Mohit Chandra, Gaurav Verma, Yibo Hu, Munmun De~Choudhury, and Srijan Kumar. 2024.
\newblock \href {https://doi.org/10.1145/3589334.3645643} {Better to ask in english: Cross-lingual evaluation of large language models for healthcare queries}.
\newblock In \emph{Proceedings of the ACM Web Conference 2024}, WWW '24, page 2627–2638, New York, NY, USA. Association for Computing Machinery.

\bibitem[{Le~Scao et~al.(2023)Le~Scao, Fan, Akiki, Pavlick, Ili{\'c}, Hesslow, Castagn{\'e}, Luccioni, Yvon, Gall{\'e} et~al.}]{le2023bloom}
Teven Le~Scao, Angela Fan, Christopher Akiki, Ellie Pavlick, Suzana Ili{\'c}, Daniel Hesslow, Roman Castagn{\'e}, Alexandra~Sasha Luccioni, Fran{\c{c}}ois Yvon, Matthias Gall{\'e}, et~al. 2023.
\newblock Bloom: A 176b-parameter open-access multilingual language model.

\bibitem[{Li et~al.(2024)Li, Chiang, Frick, Dunlap, Wu, Zhu, Gonzalez, and Stoica}]{li2024crowdsourced}
Tianle Li, Wei-Lin Chiang, Evan Frick, Lisa Dunlap, Tianhao Wu, Banghua Zhu, Joseph~E Gonzalez, and Ion Stoica. 2024.
\newblock From crowdsourced data to high-quality benchmarks: Arena-hard and benchbuilder pipeline.
\newblock \emph{arXiv preprint arXiv:2406.11939}.

\bibitem[{Luo et~al.(2024)Luo, Deng, Shen, Ng, and Chua}]{luo-etal-2024-chain}
Haohao Luo, Yang Deng, Ying Shen, See-Kiong Ng, and Tat-Seng Chua. 2024.
\newblock \href {https://doi.org/10.18653/v1/2024.acl-long.432} {Chain-of-exemplar: Enhancing distractor generation for multimodal educational question generation}.
\newblock In \emph{Proceedings of the 62nd Annual Meeting of the Association for Computational Linguistics (Volume 1: Long Papers)}, pages 7978--7993, Bangkok, Thailand. Association for Computational Linguistics.

\bibitem[{Mishra et~al.(2022)Mishra, Khashabi, Baral, and Hajishirzi}]{mishra2022cross}
Swaroop Mishra, Daniel Khashabi, Chitta Baral, and Hannaneh Hajishirzi. 2022.
\newblock Cross-task generalization via natural language crowdsourcing instructions.
\newblock In \emph{Proceedings of the 60th Annual Meeting of the Association for Computational Linguistics (Volume 1: Long Papers)}, pages 3470--3487.

\bibitem[{Muennighoff et~al.(2023)Muennighoff, Wang, Sutawika, Roberts, Biderman, Le~Scao, Bari, Shen, Yong, Schoelkopf et~al.}]{muennighoff2023crosslingual}
Niklas Muennighoff, Thomas Wang, Lintang Sutawika, Adam Roberts, Stella Biderman, Teven Le~Scao, M~Saiful Bari, Sheng Shen, Zheng~Xin Yong, Hailey Schoelkopf, et~al. 2023.
\newblock Crosslingual generalization through multitask finetuning.
\newblock In \emph{Proceedings of the 61st Annual Meeting of the Association for Computational Linguistics (Volume 1: Long Papers)}, pages 15991--16111.

\bibitem[{Qin et~al.(2024)Qin, Song, Hu, Yao, Cho, Wang, Wu, Liu, Liu, and Yu}]{qin-etal-2024-infobench}
Yiwei Qin, Kaiqiang Song, Yebowen Hu, Wenlin Yao, Sangwoo Cho, Xiaoyang Wang, Xuansheng Wu, Fei Liu, Pengfei Liu, and Dong Yu. 2024.
\newblock \href {https://doi.org/10.18653/v1/2024.findings-acl.772} {{I}n{F}o{B}ench: Evaluating instruction following ability in large language models}.
\newblock In \emph{Findings of the Association for Computational Linguistics ACL 2024}, pages 13025--13048, Bangkok, Thailand and virtual meeting. Association for Computational Linguistics.

\bibitem[{Reid et~al.(2024)Reid, Savinov, Teplyashin, Lepikhin, Lillicrap, Alayrac, Soricut, Lazaridou, Firat, Schrittwieser et~al.}]{reid2024gemini}
Machel Reid, Nikolay Savinov, Denis Teplyashin, Dmitry Lepikhin, Timothy Lillicrap, Jean-baptiste Alayrac, Radu Soricut, Angeliki Lazaridou, Orhan Firat, Julian Schrittwieser, et~al. 2024.
\newblock Gemini 1.5: Unlocking multimodal understanding across millions of tokens of context.
\newblock \emph{arXiv preprint arXiv:2403.05530}.

\bibitem[{Sakaguchi et~al.(2019)Sakaguchi, Bras, Bhagavatula, and Choi}]{sakaguchi2019winogrande}
Keisuke Sakaguchi, Ronan~Le Bras, Chandra Bhagavatula, and Yejin Choi. 2019.
\newblock Winogrande: An adversarial winograd schema challenge at scale.
\newblock \emph{arXiv preprint arXiv:1907.10641}.

\bibitem[{Sanh et~al.(2022)Sanh, Webson, Raffel, Bach, Sutawika, Alyafeai, Chaffin, Stiegler, Le~Scao, Raja et~al.}]{sanh2022multitask}
Victor Sanh, Albert Webson, Colin Raffel, Stephen~H Bach, Lintang Sutawika, Zaid Alyafeai, Antoine Chaffin, Arnaud Stiegler, Teven Le~Scao, Arun Raja, et~al. 2022.
\newblock Multitask prompted training enables zero-shot task generalization.
\newblock In \emph{ICLR 2022-Tenth International Conference on Learning Representations}.

\bibitem[{Sn{\ae}bjarnarson et~al.(2023)Sn{\ae}bjarnarson, Simonsen, Glava{\v{s}}, and Vuli{\'c}}]{snaebjarnarson-etal-2023-transfer}
V{\'e}steinn Sn{\ae}bjarnarson, Annika Simonsen, Goran Glava{\v{s}}, and Ivan Vuli{\'c}. 2023.
\newblock \href {https://aclanthology.org/2023.nodalida-1.74} {Transfer to a low-resource language via close relatives: The case study on {F}aroese}.
\newblock In \emph{Proceedings of the 24th Nordic Conference on Computational Linguistics (NoDaLiDa)}, pages 728--737, T{\'o}rshavn, Faroe Islands. University of Tartu Library.

\bibitem[{Tian et~al.(2024)Tian, Gan, Song, Zhang, and Zhang}]{tian-etal-2024-chimed}
Yuanhe Tian, Ruyi Gan, Yan Song, Jiaxing Zhang, and Yongdong Zhang. 2024.
\newblock \href {https://doi.org/10.18653/v1/2024.acl-long.386} {{C}hi{M}ed-{GPT}: A {C}hinese medical large language model with full training regime and better alignment to human preferences}.
\newblock In \emph{Proceedings of the 62nd Annual Meeting of the Association for Computational Linguistics (Volume 1: Long Papers)}, pages 7156--7173, Bangkok, Thailand. Association for Computational Linguistics.

\bibitem[{Tong et~al.(2024)Tong, Li, Wang, Wang, Teng, and Shang}]{tong-etal-2024-llms}
Yongqi Tong, Dawei Li, Sizhe Wang, Yujia Wang, Fei Teng, and Jingbo Shang. 2024.
\newblock \href {https://doi.org/10.18653/v1/2024.acl-long.169} {Can {LLM}s learn from previous mistakes? investigating {LLM}s{'} errors to boost for reasoning}.
\newblock In \emph{Proceedings of the 62nd Annual Meeting of the Association for Computational Linguistics (Volume 1: Long Papers)}, pages 3065--3080, Bangkok, Thailand. Association for Computational Linguistics.

\bibitem[{{\"U}st{\"u}n et~al.(2024){\"U}st{\"u}n, Aryabumi, Yong, Ko, D{'}souza, Onilude, Bhandari, Singh, Ooi, Kayid, Vargus, Blunsom, Longpre, Muennighoff, Fadaee, Kreutzer, and Hooker}]{ustun-etal-2024-aya}
Ahmet {\"U}st{\"u}n, Viraat Aryabumi, Zheng Yong, Wei-Yin Ko, Daniel D{'}souza, Gbemileke Onilude, Neel Bhandari, Shivalika Singh, Hui-Lee Ooi, Amr Kayid, Freddie Vargus, Phil Blunsom, Shayne Longpre, Niklas Muennighoff, Marzieh Fadaee, Julia Kreutzer, and Sara Hooker. 2024.
\newblock \href {https://doi.org/10.18653/v1/2024.acl-long.845} {Aya model: An instruction finetuned open-access multilingual language model}.
\newblock In \emph{Proceedings of the 62nd Annual Meeting of the Association for Computational Linguistics (Volume 1: Long Papers)}, pages 15894--15939, Bangkok, Thailand. Association for Computational Linguistics.

\bibitem[{Wang et~al.(2019)Wang, Pruksachatkun, Nangia, Singh, Michael, Hill, Levy, and Bowman}]{wang2019superglue}
Alex Wang, Yada Pruksachatkun, Nikita Nangia, Amanpreet Singh, Julian Michael, Felix Hill, Omer Levy, and Samuel Bowman. 2019.
\newblock Superglue: A stickier benchmark for general-purpose language understanding systems.
\newblock \emph{Advances in neural information processing systems}, 32.

\bibitem[{Wang et~al.(2018)Wang, Singh, Michael, Hill, Levy, and Bowman}]{wang-etal-2018-glue}
Alex Wang, Amanpreet Singh, Julian Michael, Felix Hill, Omer Levy, and Samuel Bowman. 2018.
\newblock \href {https://doi.org/10.18653/v1/W18-5446} {{GLUE}: A multi-task benchmark and analysis platform for natural language understanding}.
\newblock In \emph{Proceedings of the 2018 {EMNLP} Workshop {B}lackbox{NLP}: Analyzing and Interpreting Neural Networks for {NLP}}, pages 353--355, Brussels, Belgium. Association for Computational Linguistics.

\bibitem[{Wang et~al.(2024)Wang, Jiao, Huang, Dai, Huang, Tu, and Lyu}]{wang-etal-2024-countries}
Wenxuan Wang, Wenxiang Jiao, Jingyuan Huang, Ruyi Dai, Jen-tse Huang, Zhaopeng Tu, and Michael Lyu. 2024.
\newblock \href {https://doi.org/10.18653/v1/2024.acl-long.345} {Not all countries celebrate thanksgiving: On the cultural dominance in large language models}.
\newblock In \emph{Proceedings of the 62nd Annual Meeting of the Association for Computational Linguistics (Volume 1: Long Papers)}, pages 6349--6384, Bangkok, Thailand. Association for Computational Linguistics.

\bibitem[{Wei et~al.()Wei, Bosma, Zhao, Guu, Yu, Lester, Du, Dai, and Le}]{weifinetuned}
Jason Wei, Maarten Bosma, Vincent Zhao, Kelvin Guu, Adams~Wei Yu, Brian Lester, Nan Du, Andrew~M Dai, and Quoc~V Le.
\newblock Finetuned language models are zero-shot learners.
\newblock In \emph{International Conference on Learning Representations}.

\bibitem[{Wei et~al.(2023)Wei, Wei, Lin, Li, Zhang, Ren, Li, Wan, Cao, Xie et~al.}]{wei2023polylm}
Xiangpeng Wei, Haoran Wei, Huan Lin, Tianhao Li, Pei Zhang, Xingzhang Ren, Mei Li, Yu~Wan, Zhiwei Cao, Binbin Xie, et~al. 2023.
\newblock Polylm: An open source polyglot large language model.
\newblock \emph{arXiv preprint arXiv:2307.06018}.

\bibitem[{Xu et~al.(2024)Xu, Zhu, Zhao, Pan, Li, and Wang}]{xu-etal-2024-pride}
Wenda Xu, Guanglei Zhu, Xuandong Zhao, Liangming Pan, Lei Li, and William Wang. 2024.
\newblock \href {https://doi.org/10.18653/v1/2024.acl-long.826} {Pride and prejudice: {LLM} amplifies self-bias in self-refinement}.
\newblock In \emph{Proceedings of the 62nd Annual Meeting of the Association for Computational Linguistics (Volume 1: Long Papers)}, pages 15474--15492, Bangkok, Thailand. Association for Computational Linguistics.

\bibitem[{Xue et~al.(2022)Xue, Barua, Constant, Al-Rfou, Narang, Kale, Roberts, and Raffel}]{xue-etal-2022-byt5}
Linting Xue, Aditya Barua, Noah Constant, Rami Al-Rfou, Sharan Narang, Mihir Kale, Adam Roberts, and Colin Raffel. 2022.
\newblock \href {https://doi.org/10.1162/tacl_a_00461} {{B}y{T}5: Towards a token-free future with pre-trained byte-to-byte models}.
\newblock \emph{Transactions of the Association for Computational Linguistics}, 10:291--306.

\bibitem[{Yang et~al.(2024)Yang, Yang, Zhang, Hui, Zheng, Yu, Li, Liu, Huang, Wei, Lin, Yang, Tu, Zhang, Yang, Yang, Zhou, Lin, Dang, Lu, Bao, Yang, Yu, Li, Xue, Zhang, Zhu, Men, Lin, Li, Xia, Ren, Ren, Fan, Su, Zhang, Wan, Liu, Cui, Zhang, and Qiu}]{qwen2.5}
An~Yang, Baosong Yang, Beichen Zhang, Binyuan Hui, Bo~Zheng, Bowen Yu, Chengyuan Li, Dayiheng Liu, Fei Huang, Haoran Wei, Huan Lin, Jian Yang, Jianhong Tu, Jianwei Zhang, Jianxin Yang, Jiaxi Yang, Jingren Zhou, Junyang Lin, Kai Dang, Keming Lu, Keqin Bao, Kexin Yang, Le~Yu, Mei Li, Mingfeng Xue, Pei Zhang, Qin Zhu, Rui Men, Runji Lin, Tianhao Li, Tingyu Xia, Xingzhang Ren, Xuancheng Ren, Yang Fan, Yang Su, Yichang Zhang, Yu~Wan, Yuqiong Liu, Zeyu Cui, Zhenru Zhang, and Zihan Qiu. 2024.
\newblock Qwen2.5 technical report.
\newblock \emph{arXiv preprint arXiv:2412.15115}.

\bibitem[{Zellers et~al.(2019)Zellers, Holtzman, Bisk, Farhadi, and Choi}]{zellers2019hellaswag}
Rowan Zellers, Ari Holtzman, Yonatan Bisk, Ali Farhadi, and Yejin Choi. 2019.
\newblock Hellaswag: Can a machine really finish your sentence?
\newblock In \emph{Proceedings of the 57th Annual Meeting of the Association for Computational Linguistics}.

\bibitem[{Zheng et~al.(2023)Zheng, Chiang, Sheng, Zhuang, Wu, Zhuang, Lin, Li, Li, Xing et~al.}]{zheng2023judging}
Lianmin Zheng, Wei-Lin Chiang, Ying Sheng, Siyuan Zhuang, Zhanghao Wu, Yonghao Zhuang, Zi~Lin, Zhuohan Li, Dacheng Li, Eric Xing, et~al. 2023.
\newblock Judging llm-as-a-judge with mt-bench and chatbot arena.
\newblock \emph{Advances in Neural Information Processing Systems}, 36:46595--46623.

\bibitem[{Zhong et~al.(2021)Zhong, Lee, Zhang, and Klein}]{zhong2021adapting}
Ruiqi Zhong, Kristy Lee, Zheng Zhang, and Dan Klein. 2021.
\newblock Adapting language models for zero-shot learning by meta-tuning on dataset and prompt collections.
\newblock In \emph{Findings of the Association for Computational Linguistics: EMNLP 2021}, pages 2856--2878.

\bibitem[{Zhou et~al.(2023)Zhou, Lu, Mishra, Brahma, Basu, Luan, Zhou, and Hou}]{zhou2023instruction}
Jeffrey Zhou, Tianjian Lu, Swaroop Mishra, Siddhartha Brahma, Sujoy Basu, Yi~Luan, Denny Zhou, and Le~Hou. 2023.
\newblock Instruction-following evaluation for large language models.
\newblock \emph{arXiv preprint arXiv:2311.07911}.

\end{thebibliography}

\newpage
\onecolumn
\appendix

\section{Dataset}

\subsection{M-IFEval Language Specific Instructions}

\begin{table}[H]
\centering
\captionsetup{justification=centering}
\begin{tabular}{lll}
\hline
\multicolumn{1}{|l|}{\textbf{Instruction Group}} & \multicolumn{1}{l|}{\textbf{Instruction}} & \multicolumn{1}{l|}{\textbf{Description}} \\ \hline
 &  &  \\
\textbf{Spanish} &  &  \\ \hline
\multicolumn{1}{|l|}{Special Characters} & \multicolumn{1}{l|}{Letter Frequency (ñ)} & \multicolumn{1}{l|}{"ñ" should appear \{N\} times} \\ \hline
\multicolumn{1}{|l|}{Special Characters} & \multicolumn{1}{l|}{Accented Word Frequency} & \multicolumn{1}{l|}{include at least/most \{N\} words with accents} \\ \hline
\multicolumn{1}{|l|}{Special Characters} & \multicolumn{1}{l|}{Letter Frequency (ü)} & \multicolumn{1}{l|}{"ü" should appear \{N\} times} \\ \hline
\multicolumn{1}{|l|}{Punctuation} & \multicolumn{1}{l|}{Interrogation Marks} & \multicolumn{1}{l|}{Include at least one question} \\ \hline
\multicolumn{1}{|l|}{Punctuation} & \multicolumn{1}{l|}{Exclamation Marks} & \multicolumn{1}{l|}{Include at least one exclamation} \\ \hline
 &  &  \\
\textbf{French} &  &  \\ \hline
\multicolumn{1}{|l|}{Special Characters} & \multicolumn{1}{l|}{Forbidden œ and ç} & \multicolumn{1}{l|}{Do not use \{char\} characters} \\ \hline
\multicolumn{1}{|l|}{Special Characters} & \multicolumn{1}{l|}{No Accents} & \multicolumn{1}{l|}{Do not use accents} \\ \hline
\multicolumn{1}{|l|}{Special Characters} & \multicolumn{1}{l|}{Add Accents} & \multicolumn{1}{l|}{Add the correct accents to the given text} \\ \hline
\multicolumn{1}{|l|}{Detectable Content} & \multicolumn{1}{l|}{Informal Address} & \multicolumn{1}{l|}{Speak directly and informally to the user} \\ \hline
\multicolumn{1}{|l|}{Detectable Content} & \multicolumn{1}{l|}{No Digits} & \multicolumn{1}{l|}{Do not use Arabic numerals} \\ \hline
 &  &  \\
\textbf{Japanese} &  &  \\ \hline
\multicolumn{1}{|l|}{Length Constraints} & \multicolumn{1}{l|}{Number Letters} & \multicolumn{1}{l|}{Use at least/most \{N\} characters} \\ \hline
\multicolumn{1}{|l|}{Detectable Format} & \multicolumn{1}{l|}{Numbered Lists} & \multicolumn{1}{l|}{Include a numbered list of exactly \{N\} items} \\ \hline
\multicolumn{1}{|l|}{Detectable Format} & \multicolumn{1}{l|}{Taigen-dome} & \multicolumn{1}{l|}{Include exactly \{N\} taigen-dome} \\ \hline
\multicolumn{1}{|l|}{Start with / End with} & \multicolumn{1}{l|}{Unified Sentence Endings} & \multicolumn{1}{l|}{All sentences must end in \{ending\}} \\ \hline
\multicolumn{1}{|l|}{Punctuation} & \multicolumn{1}{l|}{No Periods} & \multicolumn{1}{l|}{Do not use Japanese periods} \\ \hline
\multicolumn{1}{|l|}{Script} & \multicolumn{1}{l|}{Furigana} & \multicolumn{1}{l|}{Furigana must follow all kanji} \\ \hline
\multicolumn{1}{|l|}{Script} & \multicolumn{1}{l|}{Kanji} & \multicolumn{1}{l|}{Include at least/most \{N\} kanji characters} \\ \hline
\multicolumn{1}{|l|}{Script} & \multicolumn{1}{l|}{Kansuuji} & \multicolumn{1}{l|}{All numbers must be written with Kanji} \\ \hline
\multicolumn{1}{|l|}{Script} & \multicolumn{1}{l|}{No Katakana} & \multicolumn{1}{l|}{Do not include any katakana characters} \\ \hline
\multicolumn{1}{|l|}{Script} & \multicolumn{1}{l|}{No Hiragana} & \multicolumn{1}{l|}{Do not include any hiragana characters} \\ \hline
\multicolumn{1}{|l|}{Script} & \multicolumn{1}{l|}{Katakana Only} & \multicolumn{1}{l|}{Only use katakana characters} \\ \hline
\multicolumn{1}{|l|}{Script} & \multicolumn{1}{l|}{Hiragana Only} & \multicolumn{1}{l|}{Only use hiragana characters} \\ \hline
\end{tabular}
\caption{Full list of added instructions in Spanish, French, and Japanese.}
\label{tab:all_instr}
\end{table}

\vspace{0.5cm}
\subsection{Dataset Statistics}
\label{Appendix: dataset stat}

\begin{table}[H]
\centering
\captionsetup{justification=centering}
\begin{tabular}{|l|r|r|r|r|}
\hline
\textbf{Language} & \textbf{EN} & \textbf{ES} & \textbf{FR} & \textbf{JA} \\ 
\hline
Unique Instruction Types & 25 & 30 & 30 & 33 \\ 
Total Number of Prompts & 541 & 115 & 235 & 172\\
Number of Prompts with only 1 Instructions & 305 & 100 & 146 & 128\\
Number of Prompts with 2 Instructions & 179 & 8 & 68 & 34\\
Number of Prompts with 3 Instructions & 57 & 7 & 21 & 10\\
Average Prompt Length\textsuperscript{*} & 211 & 171 & 232 & 79\\ 
Standard Deviation of Prompt Length\textsuperscript{*} & 117 &  62 & 85 & 32\\ 
\hline
\end{tabular}
\caption{Basic dataset statistics. The values reported for English (EN) represent the original IFEval dataset. \\
\textsuperscript{*}Measured in total character count, including spaces and punctuation.}
\label{tab:basic_dataset_stats}
\end{table}
\clearpage

\begin{longtable}[c]{llrrrr}
\captionsetup{justification=centering}
\hline
\multicolumn{1}{|l|}{\textbf{Instruction Group}} & \multicolumn{1}{l|}{\textbf{Instruction}} & \multicolumn{1}{r|}{\textbf{EN}} &
\multicolumn{1}{r|}{\textbf{ES}} &
\multicolumn{1}{r|}{\textbf{FR}} &
\multicolumn{1}{r|}{\textbf{JA}} \\ \hline
 &  &  \\
\textbf{Shared} &  &  \\ \hline
\multicolumn{1}{|l|}{Keywords} & \multicolumn{1}{l|}{Include Keywords} & \multicolumn{1}{r|}{39} & \multicolumn{1}{r|}{4} & \multicolumn{1}{r|}{14} & \multicolumn{1}{r|}{7} \\ \hline
\multicolumn{1}{|l|}{Keywords} & \multicolumn{1}{l|}{Keyword Frequency} & \multicolumn{1}{r|}{42} & \multicolumn{1}{r|}{4} & \multicolumn{1}{r|}{13} & \multicolumn{1}{r|}{7} \\ \hline
\multicolumn{1}{|l|}{Keywords} & \multicolumn{1}{l|}{Forbidden Words} & \multicolumn{1}{r|}{49} & \multicolumn{1}{r|}{4} & \multicolumn{1}{r|}{14} & \multicolumn{1}{r|}{7} \\ \hline
\multicolumn{1}{|l|}{Keywords} & \multicolumn{1}{l|}{Letter Frequency} & \multicolumn{1}{r|}{33} & \multicolumn{1}{r|}{4} & \multicolumn{1}{r|}{14} & \multicolumn{1}{r|}{5} \\ \hline
\multicolumn{1}{|l|}{Language} & \multicolumn{1}{l|}{Response Language} & \multicolumn{1}{r|}{31} & \multicolumn{1}{r|}{4} & \multicolumn{1}{r|}{9} & \multicolumn{1}{r|}{4} \\ \hline
\multicolumn{1}{|l|}{Length Constraints} & \multicolumn{1}{l|}{Number Paragraphs} & \multicolumn{1}{r|}{27} & \multicolumn{1}{r|}{4} & \multicolumn{1}{r|}{14} & \multicolumn{1}{r|}{7} \\ \hline
\multicolumn{1}{|l|}{Length Constraints} & \multicolumn{1}{l|}{Number Sentences} & \multicolumn{1}{r|}{52} & \multicolumn{1}{r|}{9} & \multicolumn{1}{r|}{13} & \multicolumn{1}{r|}{7} \\ \hline
\multicolumn{1}{|l|}{Length Constraints} & \multicolumn{1}{l|}{Number Words} & \multicolumn{1}{r|}{52} & \multicolumn{1}{r|}{8} & \multicolumn{1}{r|}{16} & \multicolumn{1}{r|}{-} \\ \hline
\multicolumn{1}{|l|}{Length Constraints} & \multicolumn{1}{l|}{Nth Paragraph + First Word} & \multicolumn{1}{r|}{12} & \multicolumn{1}{r|}{4} & \multicolumn{1}{r|}{11} & \multicolumn{1}{r|}{7} \\ \hline

\multicolumn{1}{|l|}{Detectable Content} & \multicolumn{1}{l|}{Postscript} & \multicolumn{1}{r|}{26} & \multicolumn{1}{r|}{4} & \multicolumn{1}{r|}{13} & \multicolumn{1}{r|}{7} \\ \hline
\multicolumn{1}{|l|}{Detectable Content} & \multicolumn{1}{l|}{Number Placeholders} & \multicolumn{1}{r|}{27} & \multicolumn{1}{r|}{4} & \multicolumn{1}{r|}{11} & \multicolumn{1}{r|}{7} \\ \hline

\multicolumn{1}{|l|}{Detectable Format} & \multicolumn{1}{l|}{Number Bullets} & \multicolumn{1}{r|}{31} & \multicolumn{1}{r|}{4} & \multicolumn{1}{r|}{11} & \multicolumn{1}{r|}{7} \\ \hline
\multicolumn{1}{|l|}{Detectable Format} & \multicolumn{1}{l|}{Title} & \multicolumn{1}{r|}{37} & \multicolumn{1}{r|}{4} & \multicolumn{1}{r|}{14} & \multicolumn{1}{r|}{7} \\ \hline
\multicolumn{1}{|l|}{Detectable Format} & \multicolumn{1}{l|}{Choose From} & \multicolumn{1}{r|}{10} & \multicolumn{1}{r|}{4} & \multicolumn{1}{r|}{8} & \multicolumn{1}{r|}{4} \\ \hline
\multicolumn{1}{|l|}{Detectable Format} & \multicolumn{1}{l|}{Minimum Number Highlighted Sections} & \multicolumn{1}{r|}{48} & \multicolumn{1}{r|}{4} & \multicolumn{1}{r|}{11} & \multicolumn{1}{r|}{7} \\ \hline
\multicolumn{1}{|l|}{Detectable Format} & \multicolumn{1}{l|}{Multiple Sections} & \multicolumn{1}{r|}{14} & \multicolumn{1}{r|}{4} & \multicolumn{1}{r|}{11} & \multicolumn{1}{r|}{7} \\ \hline
\multicolumn{1}{|l|}{Detectable Format} & \multicolumn{1}{l|}{Json Format} & \multicolumn{1}{r|}{17} & \multicolumn{1}{r|}{4} & \multicolumn{1}{r|}{8} & \multicolumn{1}{r|}{6} \\ \hline

\multicolumn{1}{|l|}{Combination} & \multicolumn{1}{l|}{Repeat Prompt} & \multicolumn{1}{r|}{41} & \multicolumn{1}{r|}{4} & \multicolumn{1}{r|}{7} & \multicolumn{1}{r|}{7} \\ \hline
\multicolumn{1}{|l|}{Combination} & \multicolumn{1}{l|}{Two Responses} & \multicolumn{1}{r|}{24} & \multicolumn{1}{r|}{4} & \multicolumn{1}{r|}{12} & \multicolumn{1}{r|}{7} \\ \hline

\multicolumn{1}{|l|}{Change Case} & \multicolumn{1}{l|}{All Uppercase} & \multicolumn{1}{r|}{25} & \multicolumn{1}{r|}{4} & \multicolumn{1}{r|}{8} & \multicolumn{1}{r|}{-} \\ \hline
\multicolumn{1}{|l|}{Change Case} & \multicolumn{1}{l|}{All Lowercase} & \multicolumn{1}{r|}{39} & \multicolumn{1}{r|}{4} & \multicolumn{1}{r|}{15} & \multicolumn{1}{r|}{-} \\ \hline
\multicolumn{1}{|l|}{Change Case} & \multicolumn{1}{l|}{Frequency of All-capital Words } & \multicolumn{1}{r|}{25} & \multicolumn{1}{r|}{8} & \multicolumn{1}{r|}{11} & \multicolumn{1}{r|}{-} \\ \hline

\multicolumn{1}{|l|}{Start with / End with} & \multicolumn{1}{l|}{End Checker} & \multicolumn{1}{r|}{26} & \multicolumn{1}{r|}{4} & \multicolumn{1}{r|}{15} & \multicolumn{1}{r|}{7} \\ \hline
\multicolumn{1}{|l|}{Start with / End with} & \multicolumn{1}{l|}{Quotation} & \multicolumn{1}{r|}{41} & \multicolumn{1}{r|}{4} & \multicolumn{1}{r|}{9} & \multicolumn{1}{r|}{7} \\ \hline
\multicolumn{1}{|l|}{Punctuation} & \multicolumn{1}{l|}{No Commas} & \multicolumn{1}{r|}{66} & \multicolumn{1}{r|}{4} & \multicolumn{1}{r|}{12} & \multicolumn{1}{r|}{7} \\ \hline

 &  &  \\
\textbf{Spanish} &  &  \\ \hline
\multicolumn{1}{|l|}{Special Character} & \multicolumn{1}{l|}{Letter Frequency (ñ)} & \multicolumn{1}{r|}{-} & \multicolumn{1}{r|}{4} & \multicolumn{1}{r|}{-} & \multicolumn{1}{r|}{-} \\ \hline
\multicolumn{1}{|l|}{Special Character} & \multicolumn{1}{l|}{Accented Word Frequency} & \multicolumn{1}{r|}{-} & \multicolumn{1}{r|}{8} & \multicolumn{1}{r|}{-} & \multicolumn{1}{r|}{-} \\ \hline
\multicolumn{1}{|l|}{Special Character} & \multicolumn{1}{l|}{Letter Frequency (ü)} & \multicolumn{1}{r|}{-} & \multicolumn{1}{r|}{4} & \multicolumn{1}{r|}{-} & \multicolumn{1}{r|}{-} \\ \hline
\multicolumn{1}{|l|}{Punctuation} & \multicolumn{1}{l|}{Interrogation Marks} & \multicolumn{1}{r|}{-} & \multicolumn{1}{r|}{4} & \multicolumn{1}{r|}{-} & \multicolumn{1}{r|}{-} \\ \hline
\multicolumn{1}{|l|}{Punctuation} & \multicolumn{1}{l|}{Exclamation Marks} & \multicolumn{1}{r|}{-} & \multicolumn{1}{r|}{4} & \multicolumn{1}{r|}{-} & \multicolumn{1}{r|}{-} \\ \hline

 &  &  \\
\textbf{French} &  &  \\ \hline
\multicolumn{1}{|l|}{Special Character} & \multicolumn{1}{l|}{Forbidden œ and ç} & \multicolumn{1}{r|}{-} & \multicolumn{1}{r|}{-} & \multicolumn{1}{r|}{11} & \multicolumn{1}{r|}{-} \\ \hline
\multicolumn{1}{|l|}{Special Character} & \multicolumn{1}{l|}{Add Accents} & \multicolumn{1}{r|}{-} & \multicolumn{1}{r|}{-} & \multicolumn{1}{r|}{7} & \multicolumn{1}{r|}{-} \\ \hline
\multicolumn{1}{|l|}{Special Character} & \multicolumn{1}{l|}{No Accents} & \multicolumn{1}{r|}{-} & \multicolumn{1}{r|}{-} & \multicolumn{1}{r|}{10} & \multicolumn{1}{r|}{-} \\ \hline
\multicolumn{1}{|l|}{Detectable Content} & \multicolumn{1}{l|}{Informal Address} & \multicolumn{1}{r|}{-} & \multicolumn{1}{r|}{-} & \multicolumn{1}{r|}{11} & \multicolumn{1}{r|}{-} \\ \hline
\multicolumn{1}{|l|}{Detectable Content} & \multicolumn{1}{l|}{No Digits} & \multicolumn{1}{r|}{-} & \multicolumn{1}{r|}{-} & \multicolumn{1}{r|}{12} & \multicolumn{1}{r|}{-} \\ \hline
 &  &  \\

\textbf{Japanese} &  &  \\ \hline
\multicolumn{1}{|l|}{Length Constraints} & \multicolumn{1}{l|}{Number Letters} & \multicolumn{1}{r|}{-} & \multicolumn{1}{r|}{-} & \multicolumn{1}{r|}{-} & \multicolumn{1}{r|}{7} \\ \hline
\multicolumn{1}{|l|}{Detectable Format} & \multicolumn{1}{l|}{Numbered Lists} & \multicolumn{1}{r|}{-} & \multicolumn{1}{r|}{-} & \multicolumn{1}{r|}{-} & \multicolumn{1}{r|}{7} \\ \hline
\multicolumn{1}{|l|}{Detectable Format} & \multicolumn{1}{l|}{Taigen-dome} & \multicolumn{1}{r|}{-} & \multicolumn{1}{r|}{-} & \multicolumn{1}{r|}{-} & \multicolumn{1}{r|}{7} \\ \hline
\multicolumn{1}{|l|}{Start with / End with} & \multicolumn{1}{l|}{Unified Sentence Endings} & \multicolumn{1}{r|}{-} & \multicolumn{1}{r|}{-} & \multicolumn{1}{r|}{-} & \multicolumn{1}{r|}{7} \\ \hline
\multicolumn{1}{|l|}{Punctuation} & \multicolumn{1}{l|}{No Periods} & \multicolumn{1}{r|}{-} & \multicolumn{1}{r|}{-} & \multicolumn{1}{r|}{-} & \multicolumn{1}{r|}{7} \\ \hline
\multicolumn{1}{|l|}{Script} & \multicolumn{1}{l|}{Furigana} & \multicolumn{1}{r|}{-} & \multicolumn{1}{r|}{-} & \multicolumn{1}{r|}{-} & \multicolumn{1}{r|}{12} \\ \hline
\multicolumn{1}{|l|}{Script} & \multicolumn{1}{l|}{Kanji} & \multicolumn{1}{r|}{-} & \multicolumn{1}{r|}{-} & \multicolumn{1}{r|}{-} & \multicolumn{1}{r|}{7} \\ \hline
\multicolumn{1}{|l|}{Script} & \multicolumn{1}{l|}{Kansuuji} & \multicolumn{1}{r|}{-} & \multicolumn{1}{r|}{-} & \multicolumn{1}{r|}{-} & \multicolumn{1}{r|}{7} \\ \hline
\multicolumn{1}{|l|}{Script} & \multicolumn{1}{l|}{No Katakana} & \multicolumn{1}{r|}{-} & \multicolumn{1}{r|}{-} & \multicolumn{1}{r|}{-} & \multicolumn{1}{r|}{7} \\ \hline
\multicolumn{1}{|l|}{Script} & \multicolumn{1}{l|}{No Hiragana} & \multicolumn{1}{r|}{-} & \multicolumn{1}{r|}{-} & \multicolumn{1}{r|}{-} & \multicolumn{1}{r|}{7} \\ \hline
\multicolumn{1}{|l|}{Script} & \multicolumn{1}{l|}{Katakana Only} & \multicolumn{1}{r|}{-} & \multicolumn{1}{r|}{-} & \multicolumn{1}{r|}{-} & \multicolumn{1}{r|}{6} \\ \hline
\multicolumn{1}{|l|}{Script} & \multicolumn{1}{l|}{Hiragana Only} & \multicolumn{1}{r|}{-} & \multicolumn{1}{r|}{-} & \multicolumn{1}{r|}{-} & \multicolumn{1}{r|}{7} \\ \hline

\caption{Number of prompts for each instruction. The values reported for English (EN) represent the original IFEval dataset.}
\label{tab:num_prompts_per_instr}
\end{longtable}

\vspace{1cm}

\begin{figure}[H]
\captionsetup{justification=centering}
  \begin{subfigure}[t]{0.48\linewidth}
    \centering
    \includegraphics[width=\linewidth]{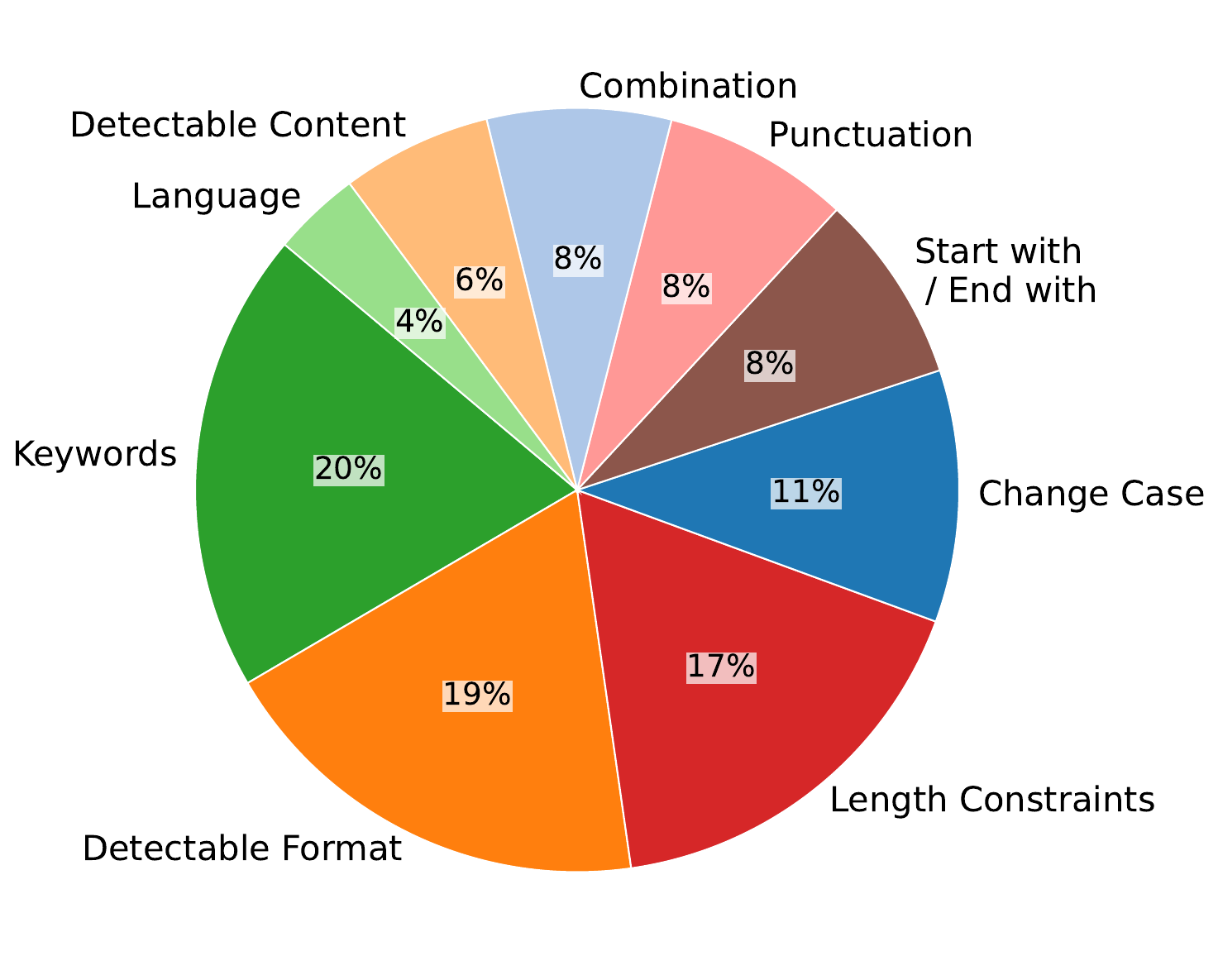}
    \caption{English (Original IFEval data)}
  \end{subfigure}
  \hfill
  \begin{subfigure}[t]{0.48\linewidth}
    \centering
    \includegraphics[width=\linewidth]{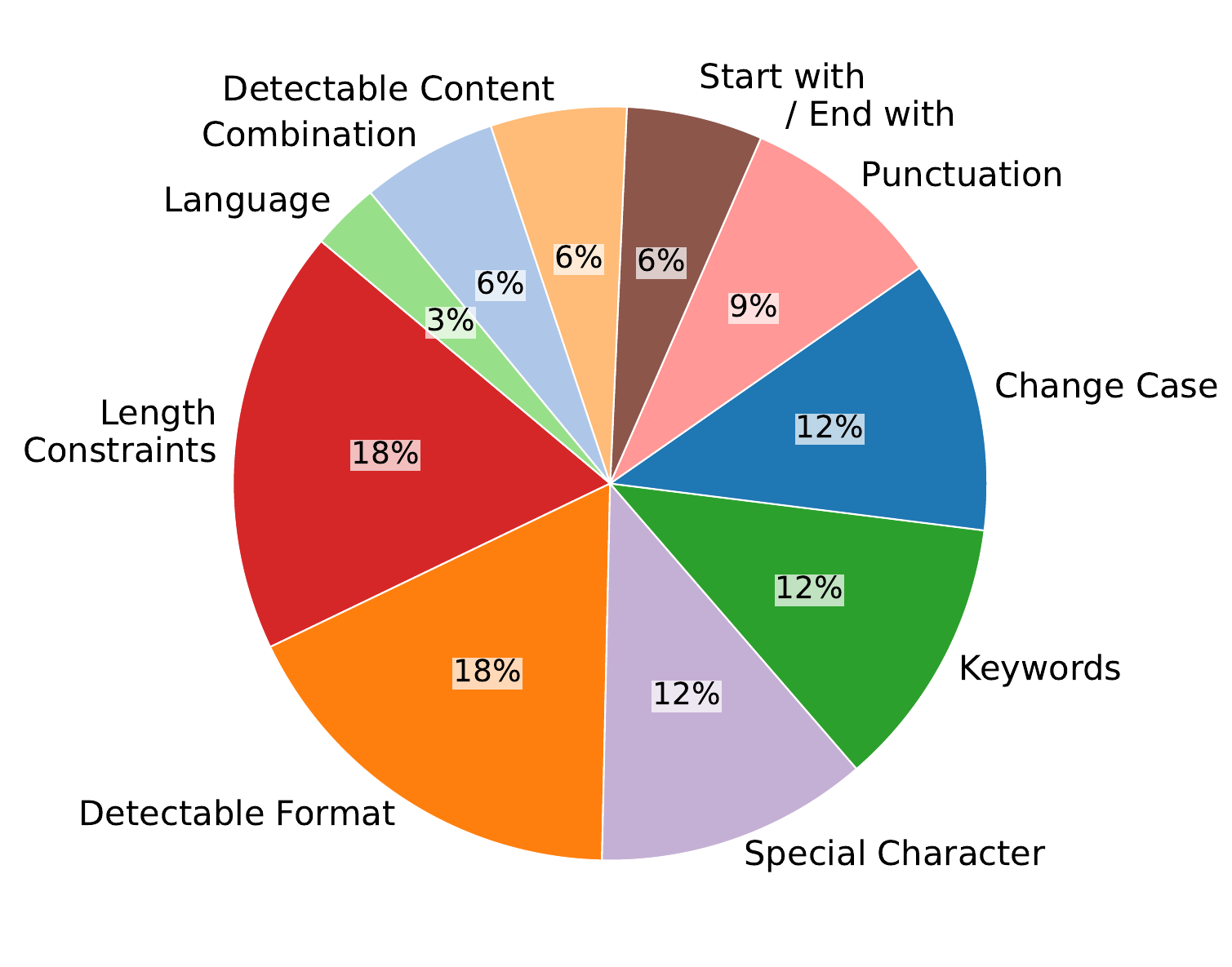}
    \caption{Spanish}
  \end{subfigure}
  \vfill
  \begin{subfigure}[t]{0.48\linewidth}
    \centering
    \includegraphics[width=\linewidth]{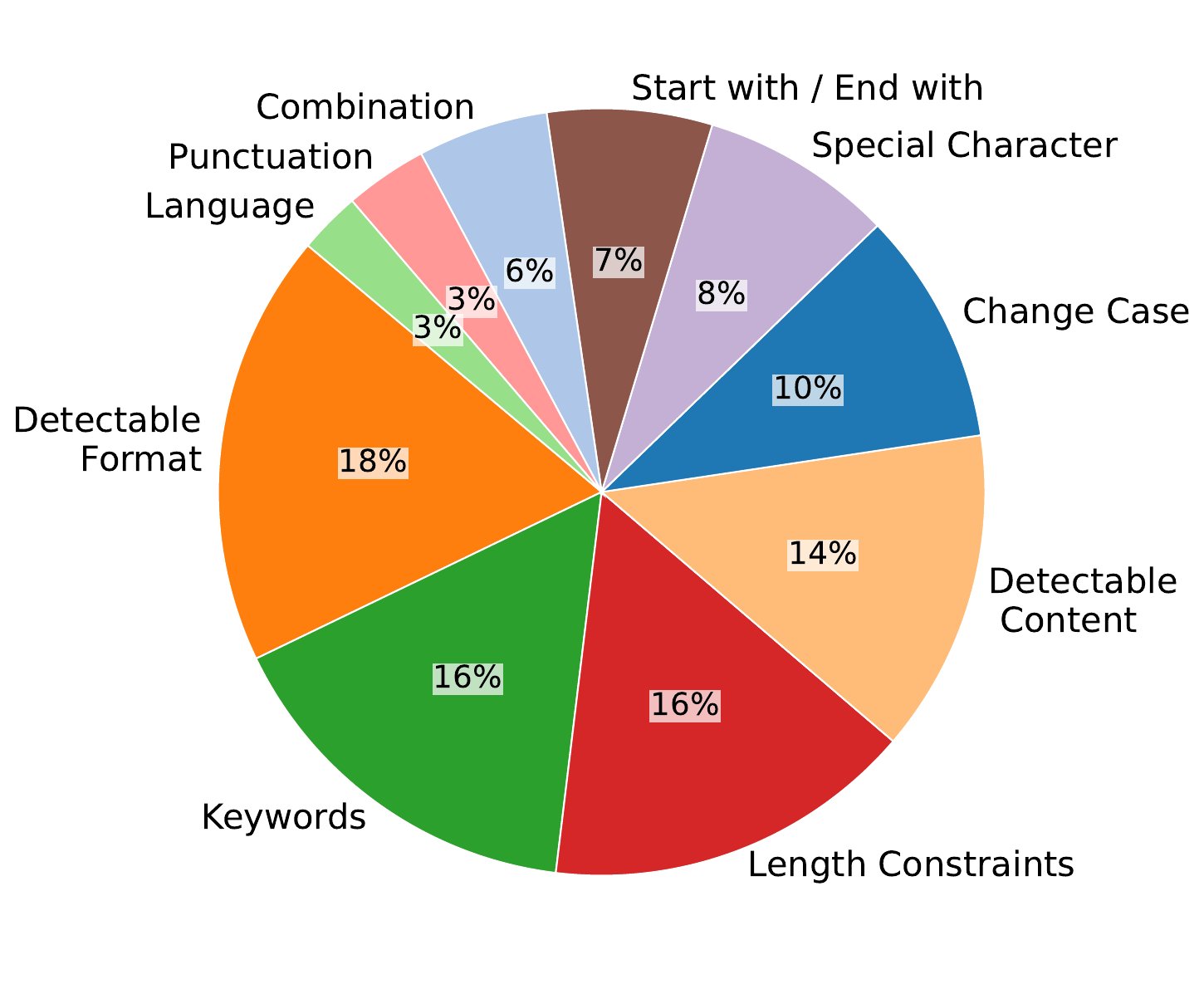}
    \caption{French}
  \end{subfigure}
  \hfill
  \begin{subfigure}[t]{0.44\linewidth}
    \centering
    \includegraphics[width=\linewidth]{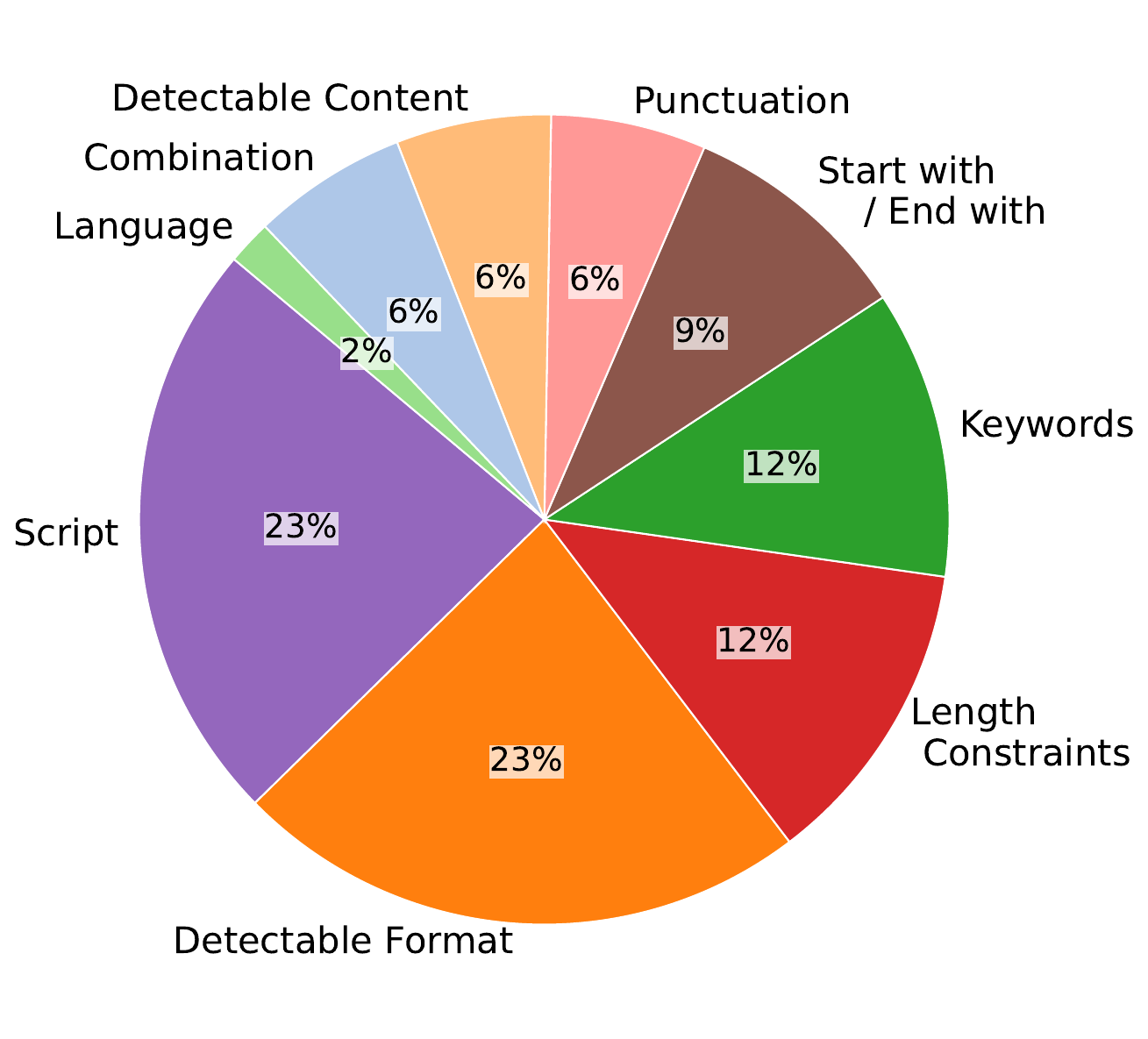}
    \caption{Japanese}
  \end{subfigure}
  \caption{Task Diversity Analysis: Percentage distribution of prompts among instruction groups by language.}
\end{figure}
\clearpage

\section{Detailed Results}
\label{Appendix: detailed results}

\begin{table}[ht]
\centering
\captionsetup{justification=centering}
\begin{tabular}{|c|rrr|r}
\cline{1-4}
\multicolumn{1}{|l|}{\textbf{Model name}} & \textbf{ES} & \textbf{FR} & \textbf{JA} & \textbf{Mean} \\ \cline{1-4}
Sonnet & 75.0 & 96.1 & \textbf{76.1} & \textbf{82.4} \\
o1 & \textbf{79.2} & \textbf{100.0} & 63.6 & 80.9 \\
Opus & 62.5 & 96.1 & 65.9 & 74.8 \\
Haiku & 58.3 & 88.2 & 59.1 & 68.6 \\
GPT4o & 58.3 & 82.4 & 63.6 & 68.1 \\
o1 Mini & 66.7 & 72.5 & 55.7 & 65.0 \\
Qwen 2.5 32B I. & 54.2 & 78.4 & 58.0 & 63.5 \\
GPT4o Mini & 58.3 & 70.6 & 58.0 & 62.3 \\ \cline{1-4}
\end{tabular}
\caption{Average loose scores of M-IFEval for each language only on the instructions that are specific to that language, sorted by the mean combined Spanish, French, and Japanese scores.}
\label{tab:avg_unique_scores_loose}
\end{table}

\begin{table}[ht]
\centering
\captionsetup{justification=centering}
\begin{tabular}{|c|r|rrr|r}
\cline{1-5}
\multicolumn{1}{|l|}{\textbf{Model name}} & \textbf{EN} & \textbf{ES} & \textbf{FR} & \textbf{JA} & \textbf{Mean} \\ \cline{1-5}
Sonnet & \textbf{93.0} & \textbf{94.9} & \textbf{94.8} & \textbf{85.0} & \textbf{91.5} \\
o1 & 89.1 & \textbf{94.9} & 93.6 & 77.4 & 88.6 \\
Opus & 92.6 & 91.2 & 92.8 & 77.9 & 87.3 \\
GPT4o & 91.2 & 92.0 & 90.4 & 76.1 & 86.2 \\
o1 Mini & 86.8 & 92.7 & 89.6 & 72.6 & 84.9 \\
GPT4o Mini & 88.7 & 89.1 & 89.3 & 71.7 & 83.3 \\
Haiku & 85.3 & 86.1 & 89.3 & 70.8 & 82.1 \\
Qwen 2.5 32B I. & 88.0 & 84.7 & 84.6 & 70.4 & 79.9 \\ \cline{1-5}
\end{tabular}
\caption{Average loose scores of M-IFEval for each language for each model evaluated, sorted by the mean combined Spanish, French, and Japanese scores.}
\label{tab:avg_scores_loose}
\end{table}


\begin{figure}[H]
    \centering
    \includegraphics[width=0.98\linewidth]{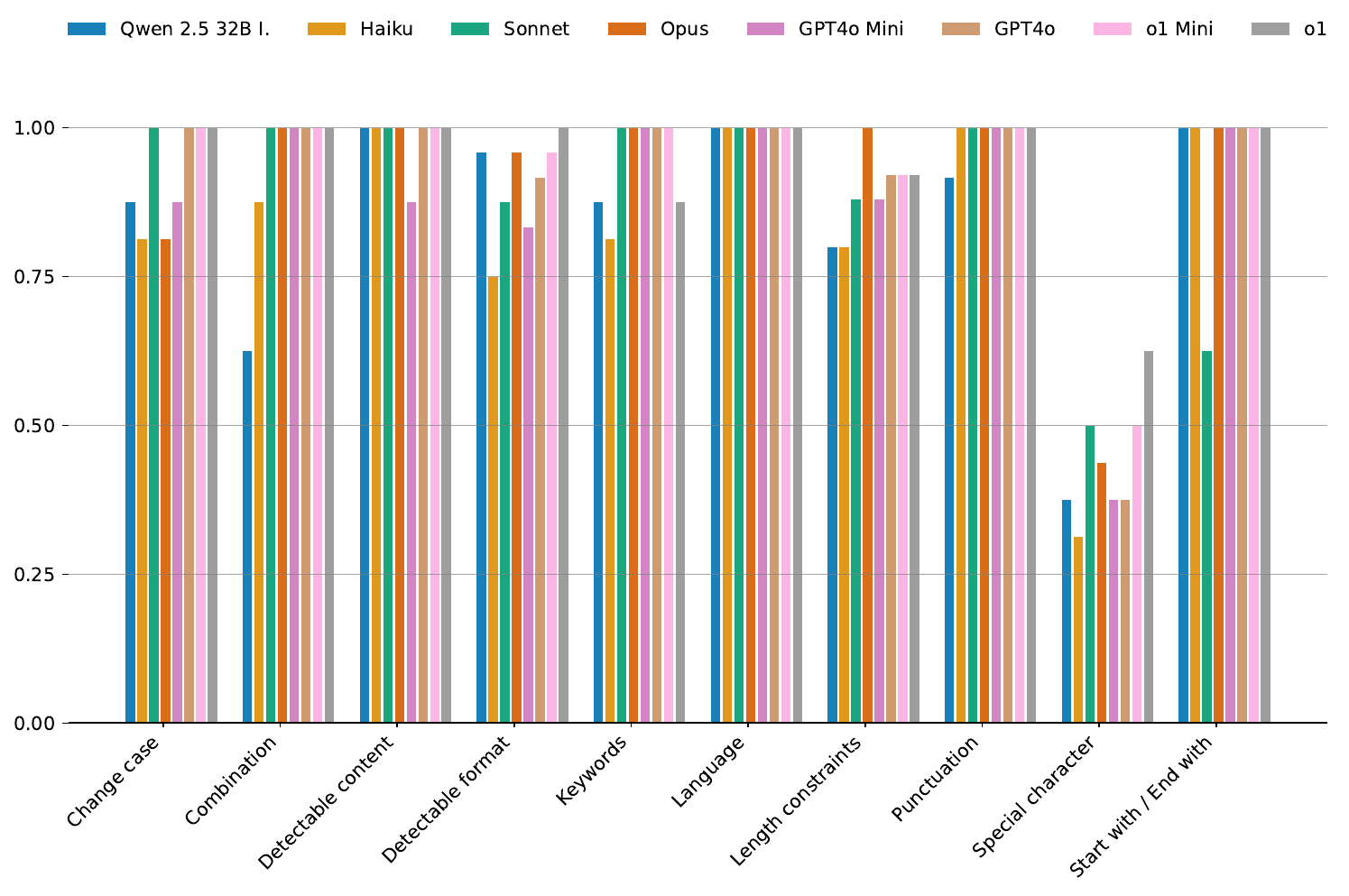}
    \caption{Instruction following strict-accuracy per instruction group: Spanish (ES).}
    \label{fig:es_performance}
\end{figure}

\begin{figure}[H]
    \centering
    \includegraphics[width=0.98\linewidth]{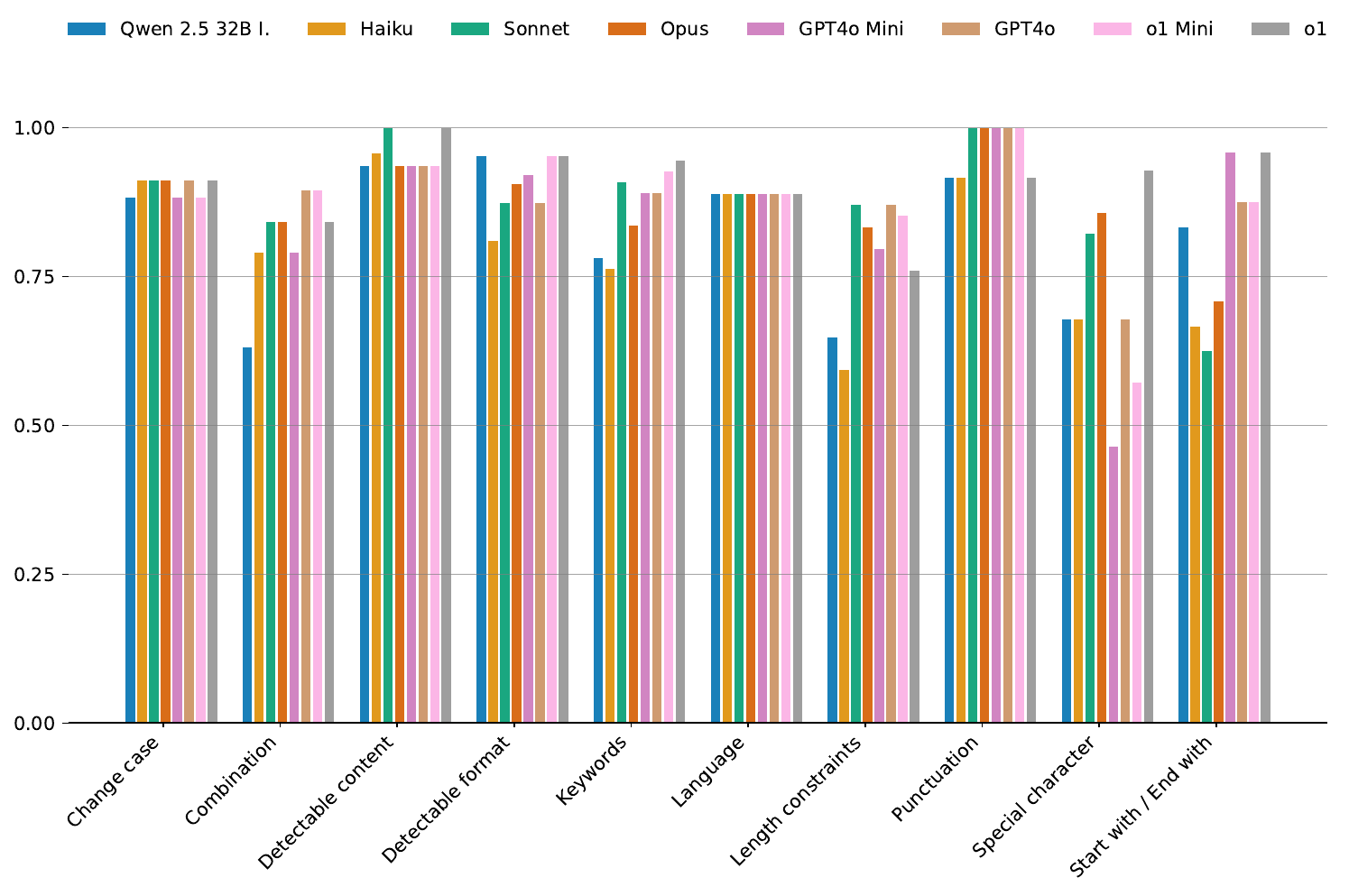}
    \caption{Instruction following strict-accuracy per instruction group: French (FR).}
    \label{fig:fr_performance}
\end{figure}

\begin{figure}[H]
    \centering
    \includegraphics[width=0.98\linewidth]{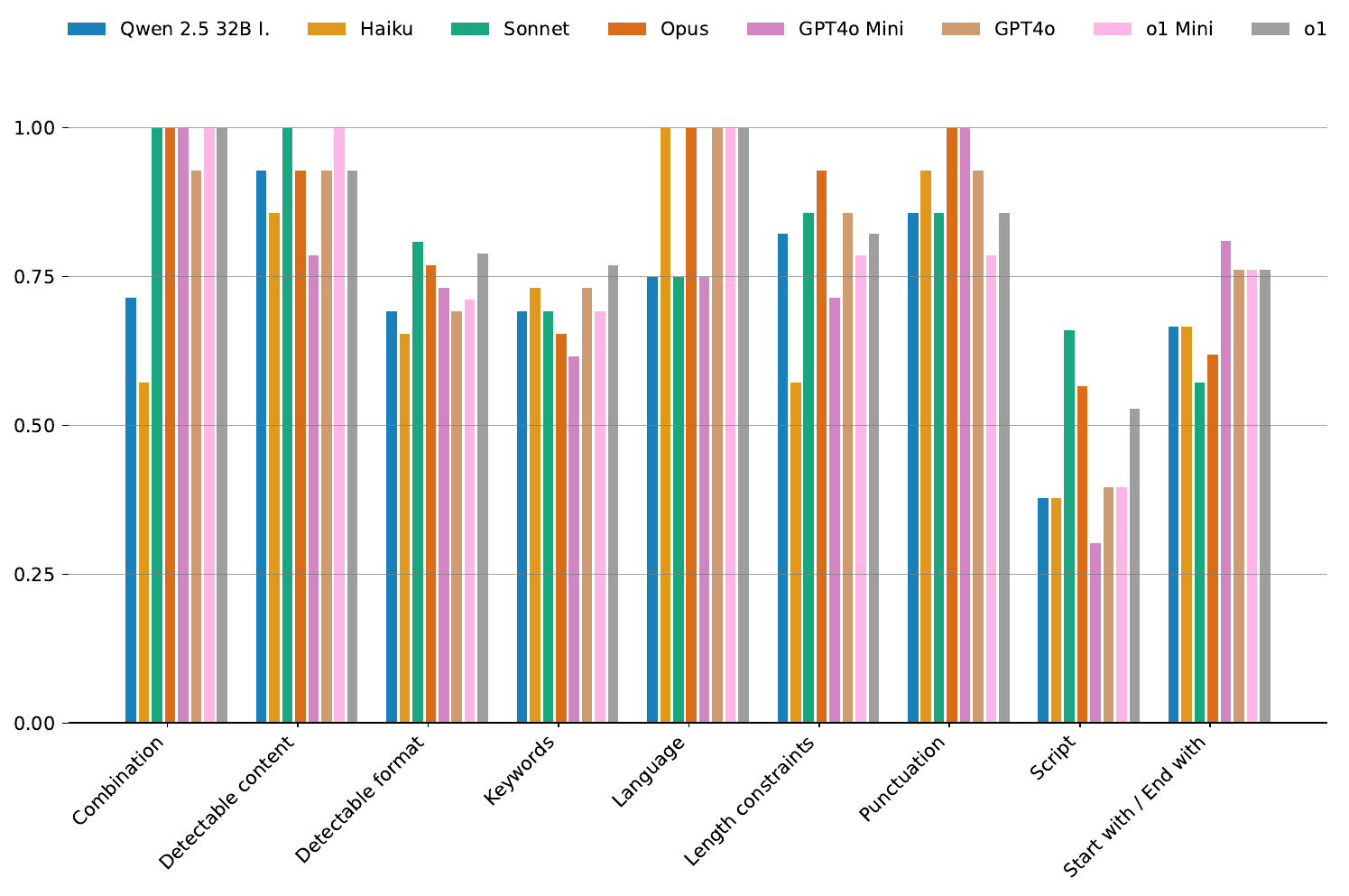}
    \caption{Instruction following strict-accuracy per instruction group: Japanese (JA).}
    \label{fig:ja_performance}
\end{figure}

\newpage
\onecolumn
\begin{longtable}[c]{|l|l|l|r|}
\captionsetup{justification=centering}
\hline
\textbf{Instruction Group} & \textbf{Instruction Name} & \textbf{Languages} & \textbf{Score} \\ \hline
\endhead
Special characters & Letter Frequency (ñ) & ES & 0.0 \\ \hline
Script & No Katakana & JA & 14.3 \\ \hline
Script & Furigana & JA & 14.6 \\ \hline
Special characters & Letter Frequency (ü) & ES & 15.6 \\ \hline
Start with / End with & Unified Sentence Endings & JA & 33.9 \\ \hline
Script & No Hiragana & JA & 35.7 \\ \hline
Script & Hiragana Only & JA & 48.2 \\ \hline
Special characters & Forbidden œ and ç & FR & 60.2 \\ \hline
Script & Katakana Only & JA & 60.4 \\ \hline
Special characters & No Accents & FR & 63.8 \\ \hline
Detectable format & Taigen-dome & JA & 64.3 \\ \hline
Detectable format & JSON Format & EN, ES, FR, JA & 67.5 \\ \hline
Length constraints & Nth Paragraph First Word & EN, ES, FR, JA & 67.6 \\ \hline
Keywords & Letter Frequency & EN, ES, FR, JA & 71.2 \\ \hline
Change cases & French Uppercase & FR & 73.4 \\ \hline
Script & Kanji & JA & 75.0 \\ \hline
Length constraints & Number Sentences & EN, ES, FR, JA & 76.4 \\ \hline
Combination & Repeat prompt & EN, ES, FR, JA & 78.2 \\ \hline
Detectable format & Number Bullets & EN, ES, FR, JA & 79.2 \\ \hline
Special characters & Accented Word Frequency & ES & 79.7 \\ \hline
Change cases & Capital Word Frequency & EN, ES, FR & 80.1 \\ \hline
Length constraints & Number Paragraphs & EN, ES, FR, JA & 80.5 \\ \hline
Length constraints & Number Words & EN, ES, FR & 82.2 \\ \hline
Keywords & Forbidden Words & EN, ES, FR, JA & 84.1 \\ \hline
Keywords & Keyword Frequency & EN, ES, FR, JA & 85.2 \\ \hline
Start with / End with & Quotation & EN, ES, FR, JA & 85.2 \\ \hline
Change cases & English Uppercase & EN & 86.0 \\ \hline
Change cases & English Uppercase & EN & 88.1 \\ \hline
Combination & Two Responses & EN, ES, FR, JA & 88.3 \\ \hline
Length constraints & Number Letters & JA & 89.3 \\ \hline
Keywords & Include Keywords & EN, ES, FR, JA & 89.6 \\ \hline
Punctuation & No Commas & EN, ES, FR, JA & 90.2 \\ \hline
Detectable content & Informal Address & FR & 90.9 \\ \hline
Letters & Kansuuji & JA & 91.1 \\ \hline
Start with / End with & End Checker & EN, ES, FR, JA & 91.3 \\ \hline
Detectable format & Title & EN, ES, FR, JA & 92.3 \\ \hline
Punctuation & No Periods & JA & 92.9 \\ \hline
Detectable content & Number Placeholder & EN, ES, FR, JA & 92.9 \\ \hline
Detectable format & Choose From & EN, ES, FR, JA & 93.3 \\ \hline
Detectable format & Minimum Number Highlighted Sections & EN, ES, FR, JA & 94.8 \\ \hline
Language & Response Language & EN, ES, FR, JA & 95.6 \\ \hline
Detectable format & Numbered Lists & JA & 96.4 \\ \hline
Punctuation & Exclamation Marks & ES & 96.9 \\ \hline
Detectable format & Multiple Sections & EN, ES, FR, JA & 97.6 \\ \hline
Detectable content & Postscript & EN, ES, FR, JA & 97.8 \\ \hline
Special characters & Add Accents & FR & 98.2 \\ \hline
Punctuation & Question Marks & ES & 100.0 \\ \hline
Change cases & Spanish Lowercase & ES & 100.0 \\ \hline
Change cases & Spanish Uppercase & ES & 100.0 \\ \hline
Change cases & French Lowercase & FR & 100.0 \\ \hline
Detectable content & No Digits & FR & 100.0 \\ \hline
\caption{The percentage strict instruction following score for each instruction, averaged over all 8 models evaluated.}
\label{tab:all_instr_averaged}\\
\end{longtable}

\section{Failure Examples}
\label{Appendix: failure examples}
Below, we present examples of failure cases in the different languages of the benchmark.

\subsection{Spanish}

\begin{tcolorbox}[colback=gray!10, colframe=gray, title=Failure Example: Special Characters - Letter Frequency (ñ)]
\textbf{Model:} o1 \hfill \textbf{Language:} Spanish (ES) 

\vspace{0.3cm}
\textbf{Prompt:} \textit{Describe el impacto de la tecnología móvil en la comunicación. En tu respuesta, las palabras que contienen la letra "ñ" deben ser 9 en total.}

\vspace{0.1cm}
\textbf{Model Response:}  
"La tecnología móvil ha transformado la comunicación en los últimos a\hl{ñ}os. Ahora, las personas pueden conectarse instantáneamente desde cualquier rincón del mundo, compartiendo experiencias y opiniones en tiempo real. Esta revolución ha desempe\hl{ñ}ado un papel esencial en la transformación de la sociedad, facilitando la interacción y el intercambio de conocimiento sin precedentes. Sin embargo, también surge la preocupación por la sobreexposición y la gestión de la privacidad. En resumen, la tecnología móvil ha ampliado nuestras oportunidades de comunicación, pero requiere una consideración consciente para equilibrar sus beneficios y desafíos."  

\vspace{0.3cm}
\textbf{Instruction (\textit{translated}):} "ñ" should appear 9 times. \faTimes

\vspace{0.1cm}
\textbf{Failure:} The response only contains 2 "ñ" when the instruction required 9.


\end{tcolorbox}

\begin{tcolorbox}[colback=gray!10, colframe=gray, title=Failure Example: Special Characters \\All Lowercase + Exclamation Marks + Accented Word Frequency]
\textbf{Model:} Qwen 2.5 32B I. \hfill \textbf{Language:} Spanish (ES) 

\vspace{0.3cm}
\textbf{Prompt:} \textit{Describe el impacto de los algoritmos de recomendación en las plataformas de streaming. Tu respuesta debe estar completamente en español, usando sólo letras minúsculas. No se permiten mayúsculas. Incluye al menos una exclamación en tu respuesta. Asegúrate de no usar más de 5 palabras con tilde en tu respuesta.}

\vspace{0.1cm}
\textbf{Model Response:}  
"los algoritmos de recomendaci\hl{ó}n transforman la experiencia en plataformas de streaming. analizan gustos, crean listas personalizadas, incrementan la retenci\hl{ó}n de usuarios. esto cambia c\hl{ó}mo exploramos contenido, nos mantiene enganchados, nos muestra lo que queremos ver antes de saberlo nosotros mismos. es fascinante c\hl{ó}mo funcionan! sin ellos, navegar\hl{í}amos ciegamente por un oc\hl{é}ano de opciones."  

\vspace{0.3cm}
\textbf{Instructions (\textit{translated}):} 
\vspace{-5pt}
\begin{itemize}
    \setlength{\itemsep}{0pt}  
    \setlength{\parskip}{0pt}
    \setlength{\topsep}{0pt}
    \setlength{\partopsep}{0pt}
    \item Answer entirely in Spanish, using only lowercase letters. \faCheck
    \item Include at least one exclamation point in your answer. \faCheck
    \item Do not use more than 5 words with an accent in your answer. \faTimes 
\end{itemize}
\vspace{-5pt}

\vspace{0.1cm}
\textbf{Failure:} The response contains 5 words with accents, one more than the instruction required.

\end{tcolorbox}

\subsection{French}

\begin{tcolorbox}[colback=gray!10, colframe=gray, title=Failure Example: Special Characters - Forbidden œ and ç]
\textbf{Model:} GPT4o \hfill \textbf{Language:} French (FR) 

\vspace{0.3cm}
\textbf{Prompt:} \textit{Quelle hormone décline pendant la ménopause, entraînant divers symptômes tels que les bouffées de chaleur ? Assurez-vous de ne pas employer de virgules dans votre texte et ne faites pas usage du graphème œ.}

\vspace{0.1cm}
\textbf{Model Response:}  
"L'hormone qui décline pendant la ménopause est l'\hl{œ}strogène."  

\vspace{0.3cm}
\textbf{Instruction (\textit{translated}):} Do not use the character "œ". \faTimes

\vspace{0.1cm}
\textbf{Failure:} The forbidden character "œ" was used once.


\end{tcolorbox}

\begin{tcolorbox}[colback=gray!10, colframe=gray, title=Failure Example: Special Characters - No Accents]
\textbf{Model:} o1 Mini \hfill \textbf{Language:} French (FR) 

\vspace{0.3cm}
\textbf{Prompt:} \textit{Concevez une rencontre entre deux figures historiques que vous respectez. Écrivez votre description sans recourir aux accents.}

\vspace{0.1cm}
\textbf{Model Response:}  
"Imaginons une rencontre entre Leonardo da Vinci et Albert Einstein dans un jardin paisible, entour\hl{é} de sculptures et de machines inventives. Leonardo, avec son esprit curieux et ses croquis toujours en main, observe attentivement les formules complexes qu’Albert \hl{é}crit sur un carnet. Les deux visionnaires \hl{é}changent des id\hl{é}es sur la nature de l'univers et les mysteres de l'energie. Leonardo partage ses dessins de machines volantes, tandis qu’Albert explique la theorie de la relativite. Leur conversation est un melange fascinant de l'art et de la science, montrant comment deux esprits brillants, bien qu’appartenant a des epoques differentes, peuvent trouver un terrain commun pour discuter et s'inspirer mutuellement. Ensemble, ils imaginent des innovations qui pourraient changer le monde, illustrant le pouvoir de la collaboration entre creativity et rationalite."  

\vspace{0.3cm}
\textbf{Instruction (\textit{translated}):} Do not use accents. \faTimes

\vspace{0.1cm}
\textbf{Failure:} Accents were used 3 times, even though that typically require accents were successfully left unaccented. 
\end{tcolorbox}

\subsection{Japanese}

\begin{tcolorbox}[colback=gray!10, colframe=gray, title=Failure Example: Script - No Katakana]
\textbf{Model:} o1 \hfill \textbf{Language:} Japanese (JA) 

\vspace{0.3cm}
\textbf{Prompt:} \begin{CJK}{UTF8}{min}
\textit{カタカナを一文字も使わないで、アフリカの伝統音楽で使用される楽器の一つを紹介してください。}
\end{CJK}

\vspace{0.1cm}
\textbf{Model Response:}  
"\begin{CJK}{UTF8}{min}\colorbox{red!30}{アフリカ}の伝統的な楽器の一つに「じぇんべ」という太鼓があります。じぇんべは西\colorbox{red!30}{アフリカ}の\colorbox{red!30}{マリ}や\colorbox{red!30}{ギニア}などで広く使われており、一枚の木をくり抜いて作られた胴に、動物の皮を張って作られます。手で叩いて演奏し、豊かな音色と多彩な\colorbox{red!30}{リズム}が特徴です。祭りや踊りの場で重要な役割を果たし、人々をつなぐ\colorbox{red!30}{コミュニケーション}の道具としても用いられています。\end{CJK}"

\vspace{0.3cm}
\textbf{Instruction (\textit{translated}):} Do not include any katakana characters. \faTimes

\vspace{0.1cm}
\textbf{Failure:} Words like \begin{CJK}{UTF8}{min}アフリカ\end{CJK} (Africa) were written in katakana, despite the existence of a non-katakana orthography: \begin{CJK}{UTF8}{min}阿弗利加\end{CJK}.
\end{tcolorbox}

\begin{tcolorbox}[colback=gray!10, colframe=gray, title=Failure Example: Script - Hiragana Only]
\textbf{Model:} Haiku \hfill \textbf{Language:} Japanese (JA) 

\vspace{0.3cm}
\textbf{Prompt:} \begin{CJK}{UTF8}{min}
\textit{ひらがなだけを使って答えてください。同じ幼稚園のけん君に告白する手紙を書いてください。}
\end{CJK}

\vspace{0.1cm}
\textbf{Model Response:}  
"\begin{CJK}{UTF8}{min}はい、わかりました。ここに、\colorbox{red!30}{同}じ\colorbox{red!30}{幼稚園}のけん\colorbox{red!30}{君}に\colorbox{red!30}{告白}する\colorbox{red!30}{手紙}を\colorbox{red!30}{書}きました。

ぼくは、けん\colorbox{red!30}{君}のことが\colorbox{red!30}{大}すきです。
いつも\colorbox{red!30}{一緒}に\colorbox{red!30}{遊}んでいて、とても\colorbox{red!30}{楽}しいです。
けん\colorbox{red!30}{君}の\colorbox{red!30}{笑顔}が、ぼくの\colorbox{red!30}{心}を\colorbox{red!30}{温}かくしてくれます。
これからも、けん\colorbox{red!30}{君}と\colorbox{red!30}{一緒}にいたいです。
ぼくの\colorbox{red!30}{気持}ちを\colorbox{red!30}{受}け\colorbox{red!30}{取}ってください。\end{CJK}"

\vspace{0.3cm}
\textbf{Instruction (\textit{translated}):} Only use hiragana characters \faTimes

\vspace{0.1cm}
\textbf{Failure:} Ignored the instruction and used kanji for most words that are typically written in kanji.

\end{tcolorbox}

\end{document}